\documentclass[lettersize,journal]{IEEEtran}
\usepackage{amsmath,amsfonts}
\usepackage{algorithmic}
\usepackage{algorithm}
\usepackage{array}
\usepackage[caption=false,font=normalsize,labelfont=sf,textfont=sf]{subfig}
\usepackage{textcomp}
\usepackage{stfloats}
\usepackage{url}
\usepackage{verbatim}
\usepackage{graphicx}
\usepackage{cite}

\usepackage{newfloat}
\usepackage{listings}
\usepackage{multirow}
\usepackage{booktabs}
\usepackage{color}
\usepackage[framemethod=tikz]{mdframed}

\usepackage[switch]{lineno}

\newcommand{\gr}[1]{\textcolor{green}{#1}}

\usepackage[hidelinks]{hyperref}
\hypersetup{
    colorlinks=true,            
    linkcolor=red,              
    citecolor=blue,            
    filecolor=mycustompurple,  
    urlcolor=magenta            
}

\begin{document}

\title{Progressive Depth Decoupling and Modulating\\ for Flexible Depth Completion}


\author{{Zhiwen Yang, Jiehua Zhang, Liang Li, Chenggang Yan, Yaoqi Sun, and Haibing Yin}
\thanks{This work was supported in part by the National Key R\&D Program of China under Grant (2023YFB4502800, 2023YFB4502803, 2020YFB1406604), National Nature Science Foundation of China (61931008, U21B2024, 62071415, 62322211, 62336008), Zhejiang Provincial Natural Science Foundation of China (LDT23F01011F01, LDT23F01015F01, LDT23F01014F01), ``Pioneer" and ``Leading Goose" R\&D Program of Zhejiang Province (2022C01068) and the Key R\&D Plan Project of Zhejiang Province (2024C01023).
}
\thanks{Corresponding author: Liang Li.}
\thanks{Zhiwen Yang is with School of Automation, Hangzhou Dianzi University, Hangzhou 310018, China (email: zhiwen.yang@hdu.edu.cn)}
\thanks{Jiehua Zhang is with School of Software Engineering, Xi'an Jiaotong University, Xi'an 614202, China (email: jiehua.zhang@stu.xjtu.edu.cn)}
\thanks{Liang Li is with the Institute of Computing Technology, Chinese Academy
of Sciences, Beijing 100089, China (e-mail: liang.li@ict.ac.cn)}
\thanks{Chenggang Yan is with School of Communication Engineering, Hangzhou Dianzi University, Hangzhou 310018, China (e-mail: cgyan@hdu.edu.cn)}
\thanks{Yaoqi Sun is with School of Communication Engineering, Hangzhou Dianzi University and Lishui Institute of Hangzhou Dianzi University, Hangzhou 310018, China (email: syq@hdu.edu.cn)}
\thanks{Haibing Yin is with School of Communication Engineering, Hangzhou Dianzi University and Lishui Institute of Hangzhou Dianzi University, Hangzhou 310018, China (e-mail: yhb@hdu.edu.cn)}}



\maketitle

\begin{abstract}
Image-guided depth completion aims at generating a dense depth map from sparse light detection and ranging (LiDAR) data and the corresponding RGB image, which is crucial for applications that require 3D scene perception, such as augmented reality and human-computer interaction.
Recent methods have shown promising performance by reformulating it as a classification problem with two sub-tasks: depth discretization and probability prediction.
They divide the depth range into several discrete depth values as depth categories, serving as priors for scene depth distributions.
However, previous depth discretization methods are easy to be impacted by depth distribution variations across different scenes, resulting in suboptimal scene depth distribution priors. 
To address the above problem, we propose a progressive depth decoupling and modulating network, which incrementally decouples the depth range into bins and adaptively generates multi-scale dense depth maps in multiple stages. Specifically, we first design a Bins Initializing Module (BIM) to construct the seed bins by exploring the depth distribution information within a sparse depth map, adapting variations of depth distribution. 
Then, we devise an incremental depth decoupling branch to progressively refine the depth distribution information from global to local. 
Meanwhile, an adaptive depth modulating branch is developed to progressively improve the probability representation from coarse-grained to fine-grained.
And the bi-directional information interactions are proposed to strengthen the information interaction between those two branches~(sub-tasks) for promoting information complementation in each branch. 
Further, we introduce a multi-scale supervision mechanism to learn the depth distribution information in latent features and enhance the adaptation capability across different scenes.
Experimental results on public datasets demonstrate that our method outperforms the state-of-the-art methods. We will release the source codes and pretrained models.
\end{abstract}


\section{Introduction}
\IEEEPARstart{D}{epth} perception is critical for many practical applications, such as robotics, autonomous driving, and augmented reality. Benefiting from the cost-effective characteristic of monocular cameras, monocular depth estimation~\cite{basak2020monocular, yang2023raildepth, shen2023dna, liu2023real, patil2022p3depth} has become one of the popular methods of acquiring depth information. 
Nevertheless, suffering from the inherent ambiguity of the mapping from 2D to 3D space, monocular systems usually generate low-precision depth maps. To solve this problem, researchers introduce extremely sparse depth maps from depth sensors as a depth calibration and transform the monocular depth estimation into depth completion task~\cite{ma2018sparse, cui2022dense}.
In virtue of the depth guidance in sparse depth maps, the mapping range from color space to depth space is restricted, and the accuracy of depth prediction is significantly advanced.

\begin{figure}[!tp] 
    \centering
    \includegraphics[width=\columnwidth]{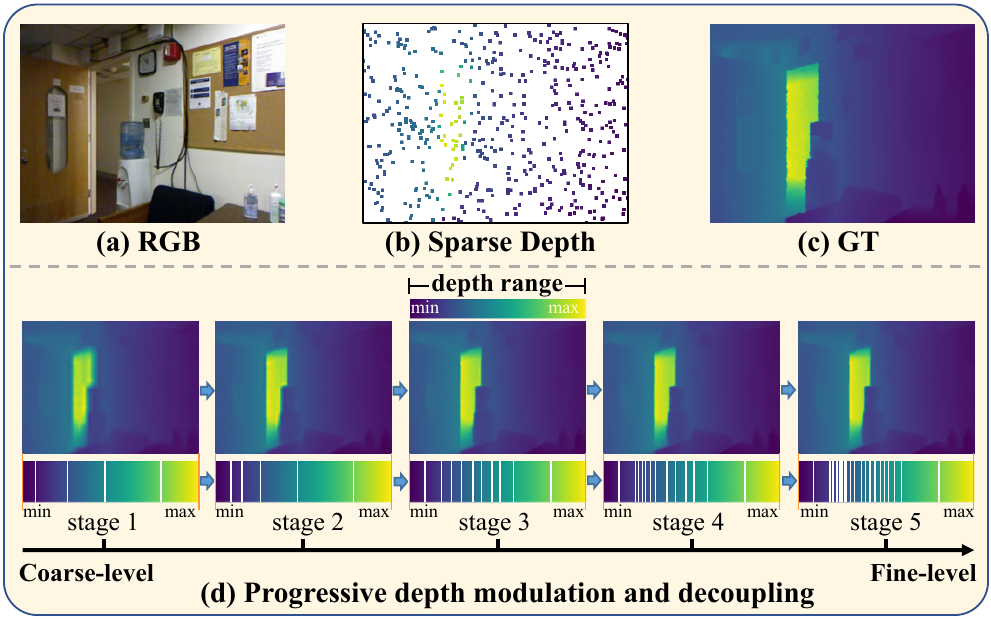}
    \caption{Illustration of our progressive depth decoupling and modulating method. \textbf{Top:} the input of the model and the ground truth. \textbf{Bottom:} an overview of the multi-scale modulated depth maps and decoupled bin partitions. Particularly, the ``min” and ``max” indicate the depth boundary of the dataset.
    }
    \label{fig1}
    \vspace{-10pt}
\end{figure}

In general, previous researchers usually solve depth completion problem through regression-based approaches which directly predict the depth value of each pixel. Some works~\cite{fu2020depth, tang2020learning, liu2021fcfr, li2020multi, cheng2018depth, cheng2020cspn++, lin2022dynamic, zhao2021adaptive} explore the guidance of geometry information within sparse depth map. They leverage geometric constraints, structured models, and graph-based networks to fill in missing depth information.
Others~\cite{zhang2018deep, qiu2019deeplidar, xu2019depth, jaritz2018sparse, zhang2021multitask} explore the complementary information between different modalities. They improve the accuracy and robustness of the network by joint training of multi-modal data or utilizing the correlation between multi-modal data.
Although these methods achieve astonishing results, they still have to predict depth values in an unbounded range, which makes these algorithms hard to converge. Recently, some methods~\cite{lee2021depth, kam2022costdcnet} show promising performance in tackling the aforementioned challenge by introducing depth range priors. They seek to model the depth completion task as a classification problem that consists of two sub-tasks, namely depth discretization and probability prediction. The former divides the depth range into several discrete depth values as depth categories, and the latter aims at producing probabilities of pixels corresponding to each depth category.
Such a classification paradigm represents the depth of each pixel as a linear combination of depth categories, which are weighted by probability predictions. That is to say, the classification-based depth completion methods heavily rely on accurate depth category generations for predicting precise depth maps. To address this challenge, researchers devote numerous efforts to exploring various depth discretization methods for accurate depth categories.

Some works~\cite{lee2021depth, fu2018deep, kam2022costdcnet} devise hand-crafted discretization strategies where depth categories remain fixed across different scenes, Lee \textit{et al.}~\cite{lee2021depth} divide the depth range into equally spaced intervals and predict a calibration bias to refine the depth prediction. However, the uniform depth categories cannot match the depth distributions in different scenes. Therefore, to adapt the scene depth distribution variation, some~\cite{bhat2021adabins, peng2022pixelwise} propose to dynamically discretize the depth range based on the depth information of scenes by an extra neural network. For instance, work~\cite{peng2022pixelwise} generates tailored bin partition for each pixel based on a lightweight auxiliary branch which extracts geometric features from sparse depth maps. Nevertheless, they directly perform depth discretization with a single-scale feature (\textit{i.e.} the highest resolution feature map from the end of the network), inevitably leading to the loss of crucial global information regarding the depth distribution representation in the scene, resulting in inaccurate depth category predictions. Therefore, to address the difficulty of generating precise depth categories, we devise a progressive refinement strategy that initially utilizes global information to generate coarse depth categories, and progressively integrates local information to achieve fine-grained depth categories aligned with the depth distribution of the scene.

This paper proposes a progressive depth decoupling and modulating network for flexible depth completion, which incrementally decouples the depth range into bins and adaptively generates multi-scale dense depth maps in a coarse-to-fine manner as illustrated in Fig.~\ref{fig1}. For the depth decoupling branch, we first design a lightweight bins initializing module to transform the spatial coordinate information within the sparse depth map into seed bins, which reflect the complete scene depth distribution, as illustrated in Fig.~\ref{fig3}. Then, the seed bins are incrementally refined from global to local by the cross-attention with the depth feature, which is generated from the depth modulating branch. Such a progressive matching between bin embedding and depth feature from image patch to pixel level enables bin partition to focus on depth intervals containing more pixels. Further, each bin embedding is transformed to the depth feature space to guide the depth modulating, which learns a compact depth feature and the corresponding probability prediction. For the depth modulating branch, we first generate area-level coarse probability prediction. Sequentially, the probability prediction is adaptively refined by assigning larger weights to bin partitions close to the pixel as the information of depth features increases. Additionally, benefitting from our progressive modulating strategy, we can flexibly predict depth maps of different scales and accuracies. Finally, we design a multi-scale loss that performs depth supervision at each incremental stage for matching the bin embedding with the depth feature and generates intermediate guidance with precise geometric information for subsequent stages. Thus, the progressive accurate depth categories and probability predictions contribute to a more precise final-scale depth map.

The main contribution can be summarized as follows:  
\begin{itemize}
\item We propose a progressive depth decoupling and modulating framework, which mitigates the difficulty of directly generating accurate depth categories and tackles the problem that classification-based depth completion methods fail to obtain precise depth categories.
\item We devise the bins initialization module which addresses the issue of previous depth
discretization methods being susceptible to variations in depth distribution by exploring the depth distribution prior in sparse depth maps.
\item Extensive experiments on public datasets show our method achieves superior depth completion performance. Visualization results also indicate our method can capture the depth information of tiny objects.
\end{itemize}

\section{Related Work}
\subsection{Depth Completion}
Depth completion is an important and challenging task that aims to generate a dense depth map through a sparse depth map. Early methods~\cite{uhrig2017sparsity, chodosh2019deep, eldesokey2018propagating} only leveraged the limited information within sparse depth maps for depth completion. Then, to overcome the lack of semantic information, researchers introduced other modalities such as event images~\cite{cui2022dense} and RGB images~\cite{ma2018sparse}. Subsequently, the focus of research shifted towards image-guided depth completion due to the accessible properties of RGB images.
The pioneer work~\cite{ma2018sparse} introduced RGB images to guide depth completion and boosted the accuracy with a large margin. Then, researchers subsequently invented various technologies to generate more precision predictions. Works~\cite{fu2020depth, tang2020learning, zhong2019deep, yan2022rignet} adopt a multi-branch fusion manner for the complementation between image feature and depth feature. Works~\cite{liu2021fcfr, hegde2021deepdnet, liu2021learning, li2020multi, tang2020learning} predicted the depth map from coarse-level to fine-level. Works~\cite{zhang2018deep, qiu2019deeplidar, xu2019depth} introduced the constraint between surface normal map and depth map to promote prediction accuracy. Works~\cite{cheng2018depth, cheng2020cspn++, xu2020deformable, lin2022dynamic, hu2021penet, Zhang2023CompletionFormer} iteratively propagated the depth of surrounding pixels based on the affinity kernels by a spatial propagation network (SPN). These methods treated depth completion as a regression problem and directly regressed the depth of each pixel. Recently, Lee \textit{et al.}~\cite{lee2021depth} reformulated the depth completion task into a combination of depth plane classification and residual regression. They first discretized the depth space into fixed planes. Then, they predicted the depth plane belonging to each pixel and represented depth as the combination of the depth of these planes plus a residual value. Kam \textit{et al.}~\cite{kam2022costdcnet} leveraged designed fixed planes to form a feature volume based on both 2D and 3D features and form a cost volume through 3D encoder for depth probability estimation. Different from the hand-crafted discretization strategy, Peng \textit{et al.}~\cite{peng2022pixelwise} proposed a pixel-wise adaptive discretization method that generates the tailored bin partition for each pixel. Existing methods focus on fixing inaccuracies of the generated depth categories through various algorithms. In contrast, our approach incrementally generates the final accurate depth categories from coarse to fine by progressive depth decoupling and modulating strategy.

\begin{figure*}[!ht]
    \centering
    \includegraphics[scale=0.65]{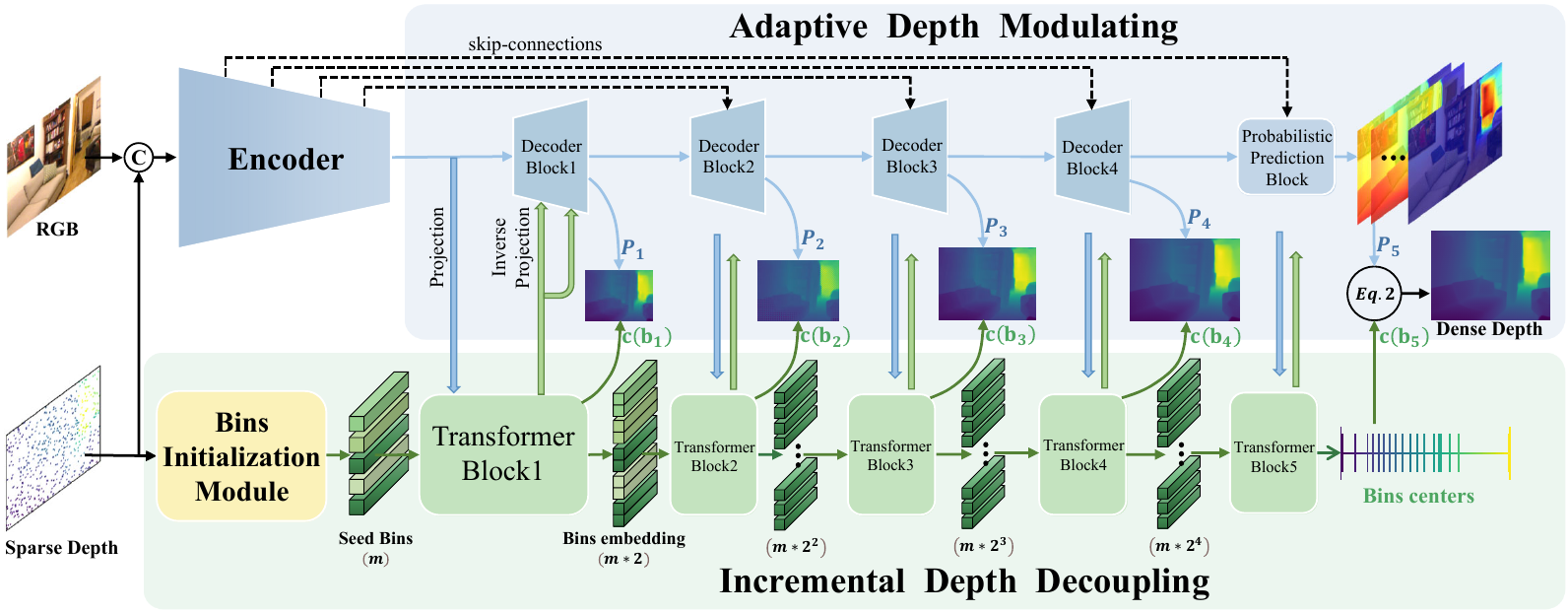}
    \caption{Overview of the proposed network architecture, which consists of an encoder, a depth modulating branch, and a depth decoupling branch. 
    The input to network is an RGB image and a sparse depth map, and the output of each stage is a dense depth map produced by the result of these two branches with Eq.~\ref{depth predict}.
    }
    \label{fig2}
\vspace{-10pt}
\end{figure*}

\subsection{Depth Discretization}
Our method is also related to some monocular depth estimation methods which estimate depth values from a single image in a classification paradigm. As a pioneer work, DORN~\cite{fu2018deep} formulated the depth estimation as an ordinal regression problem and achieved impressive performance. Nevertheless, their depth discretization is fixed and independent of scene changes. To guide network focus on the region where depth values are more frequently occurring, Adabins~\cite{bhat2021adabins} proposed to adaptively generate bin partition through a mini-ViT module with the feature from the final layer of the network. However, Binsformer~\cite{li2022binsformer} argued that the limited receptive fields of the last decoder block's feature lead to an inevitable loss of global information for depth discretization. Therefore, they introduced decoder features of all scales for guiding depth discretizing.
Meanwhile, to solve the problem of late injection in Adabins~\cite{bhat2021adabins}, Localbins~\cite{bhat2022localbins} gradually refined the bin partitions for each pixel during decoding stages. Existing works address how to utilize semantic information within RGB images to guide depth category generation. In contrast, leveraging the characteristic of sparse depth maps to represent the spatial distribution of the entire scene, we design a simple and effective Bins Initialization Module to capture the depth distribution prior from sparse depth maps for better depth category generation.

\section{Methodology}
In this section, we overview the paradigm of classification-based depth completion method and the architecture of our network. Then we introduce our lightweight bins initializing module. Next, we detail our progressive depth decoupling and modulating strategy respectively. Finally, we discuss the loss functions for multi-scale supervision. 

\subsection{Overview}
Classification-based depth completion paradigm usually involves two sub-tasks named depth discretization and probability prediction~\cite{lee2021depth}. Depth discretization, such as uniform depth discretization, divides the continuous depth range $[d_{min}, d_{max}]$ into a set of ordered bins which are also known as bin partition $\textbf{b}=[b_1, b_2, \cdots, b_N]\in \mathbb{R}^N $, where $N$ means the number of bins. Then the depth of each bin $b_i$ is represented by the bin center $c(b_i)$:
\begin{equation}\label{bins center}
    c\left(b_i\right)=d_{\min }+\left(d_{\max }-d_{\min }\right)\left(\frac{b_i}{2}+\sum_{j=1}^{i-1} b_j\right).
\end{equation}

\begin{figure}[!t]
    \centering
    \includegraphics[width=0.95\columnwidth]{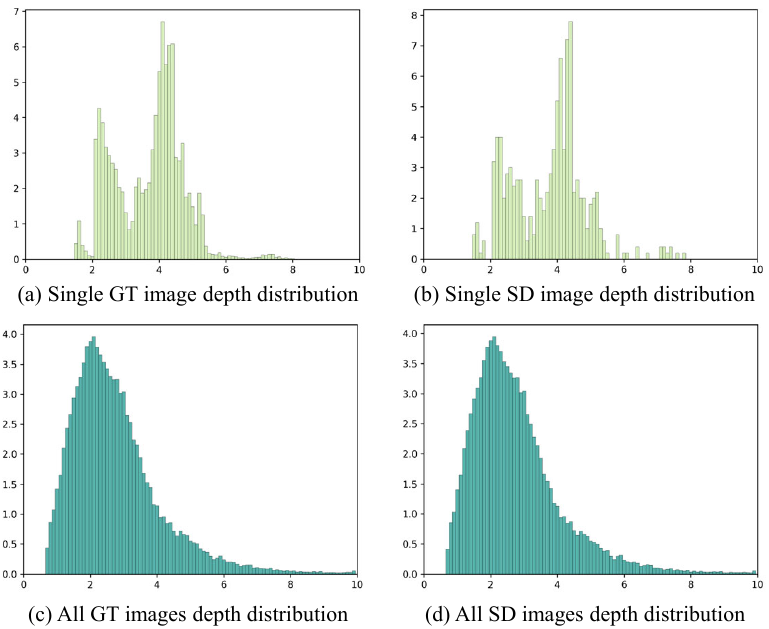}
    \caption{The comparison of the depth distribution between ground truth and sparse depth map. (a) and (b) show the depth distribution of a single image. (c) and (d) represent the depth distribution in the whole NYU-Depth-v2 dataset. The horizontal axis denotes the depth value, and the vertical axis is the percentage of pixels for each depth value out of the total number of pixels.
    }
    \label{fig3}
    \vspace{-10pt}
\end{figure}

Another sub-task estimates the possibilities of each pixel belonging to different bins. 
The predicted probability vector $\textbf{p}=[p_1, p_2, \cdots, p_N]\in \mathbb{R}^N$ denotes a pixel's distribution across bins. 
Finally, the depth $\hat{d}$ of each pixel is represented as a linear combination of bin centers $\textbf{c}(\textbf{b})$, which are weighted by the probability prediction $\textbf{p}$:
\begin{equation}\label{depth predict}
    \hat{d}=\sum_{i=1}^{N} p_i c\left(b_i\right).
\end{equation}
Targeting at such two sub-tasks, our network mainly comprises two branches, including depth decoupling and depth modulating, as illustrated in Fig.~\ref{fig2}. Moreover, the projection and inverse projection routes are designed to promote cross-branch interaction. The depth decoupling branch consists of a bins initializing module and five transformer blocks which decouple the depth range into bins with the depth distribution prior from sparse depth maps. Concretely, the bins initializing module transforms a sparse depth map into seed bins which involve the overall scene depth distribution information. And the seed bins sequentially fuse depth features from the depth modulating branch through transformer blocks to produce bin embeddings and further generate the bin centers at each stage. Meanwhile, the bin embeddings are expanded to involve more detailed depth distribution information. The depth modulating branch, which is composed of an encoder, four decoder blocks, a probabilities prediction block, and a CSPN module. It takes an RGB image and a sparse depth map as input and progressively predicts probability representation.
Specifically, we adapt ResNet34 as the backbone to extract five scaled features. And the decoder block is mainly implemented by transpose convolution layers. It receives depth features from the previous layer, the bin embedding from the depth decoupling branch, and the skipped feature from the encoder layer. Then these decoder blocks output a modulated depth feature and a probability representation corresponding to each bin partition. Such bi-directional information interactions ensure mutual guidance between these two branches. Next, our network generates multi-scale depth maps by combining the bin centers and the probability representation at each stage, which is supervised by multi-scale loss. 
Finally, a CSPN module~\cite{cheng2018depth} is introduced to refine the final depth map.

\subsection{Bins Initialization Module}
Due to the lack of explicit position guidance, previous approaches~\cite{lee2021depth, kam2022costdcnet, peng2022pixelwise} inefficiently captured depth distribution. Here, we reveal that the depth distribution of different scenarios can be represented by the random sampled sparse depth map, as illustrated in Fig.~\ref{fig3}. Moreover, depth discretization relies on the scene depth distribution which is linked to the spatial coordinates of each valid pixel. Thus, we introduce the spatial information within sparse depth maps to guide the depth discretization.

\begin{figure}[t]
    \centering
    \includegraphics[width=1\columnwidth]{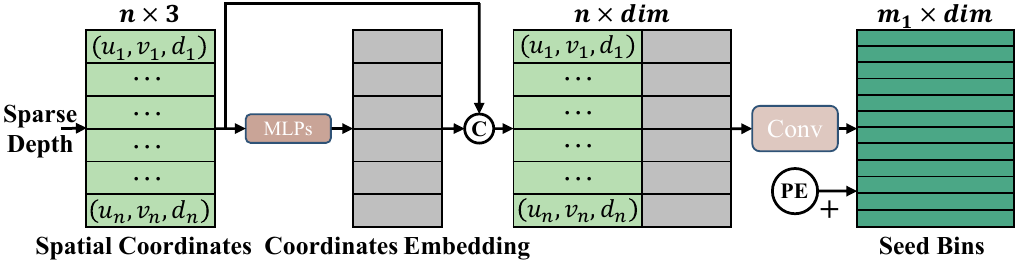}
    \caption{An overview of our BIM architecture, which takes sparse depth map as input and generates the seed bins.
    }
    \label{fig4}
\end{figure}

A direct method of exploiting spatial information is to extract features through convolution neural networks.
Nevertheless, traditional grid-based convolution is inefficient in extracting valid information from sparse inputs abundant with zero values. For instance, the sparse depth map measured by the 64-beam LiDARs only has about $2\%$ valid depth reference points. To leverage the depth distribution information within the sparse depth map more effectively, inspired by the work of point cloud analysis~\cite{zhang2023flattening}, we design the lightweight Bins Initializing Module (BIM) for point-wise mappings between 2D and 3D space, as shown in Fig.~\ref{fig4}. Specifically, we first extract horizontal and vertical coordinates $(\textbf{u}_n, \textbf{v}_n)$ in the image coordinate system of those $n$ valid sample points in the sparse depth map. Then we combine them together with corresponding depth values $d_n$ to form a triple $(\textbf{u}_n, \textbf{v}_n, \textbf{d}_n)$ as spatial coordinates $\boldsymbol{S} \in \mathbb{R}^{n \times 3}$. To maintain the flexibility of our method, we use image coordinate system instead of world coordinates as it relies on the specific camera intrinsic parameters. Subsequently, these spatial coordinates $\boldsymbol{S}$ are fed into multi-layer perceptions ($MLPs$) to learn high-dimension spatial distribution feature $\boldsymbol{C} \in \mathbb{R}^{n \times dim}$ as coordinates embedding, where $dim$ is the embedding dimension:
\begin{equation}\label{coordicates embedding}
    \boldsymbol{C} = concat\left(\boldsymbol{S}, MLPs(\boldsymbol{S})\right).
\end{equation}
Finally, we pass the coordinates embedding through a convolution layer to squeeze the channel into the initial bin number $m_1$ and get the seed bins $\boldsymbol{B}_{seed} \in \mathbb{R}^{m_1 \times dim}$ add with a learnable position embedding $\boldsymbol{PE} \in \mathbb{R}^{m_1 \times dim}$:
\begin{equation}\label{bins query}
    \boldsymbol{B}_{seed} = Conv_{1d}(Reshape(\boldsymbol{C})) + \boldsymbol{PE}.
\end{equation}

\begin{figure}[b]
    \centering
    \includegraphics[width=1\columnwidth]{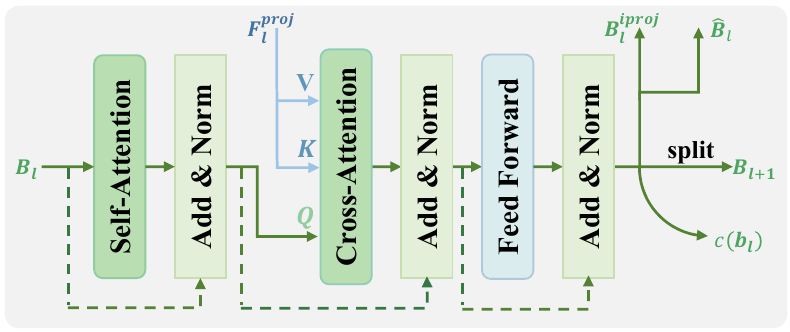}
    \caption{
    An overview of our transformer block architecture, which receives bin embedding and the projected depth feature as input, and the result is sent to three paths to promote depth modulating, refine bin embedding, and generate bin centers.
    }
    \label{fig:trans}
\end{figure}

\subsection{Incremental Depth Decoupling}
\begin{figure}[t]
    \centering
    \includegraphics[width=0.75\columnwidth]{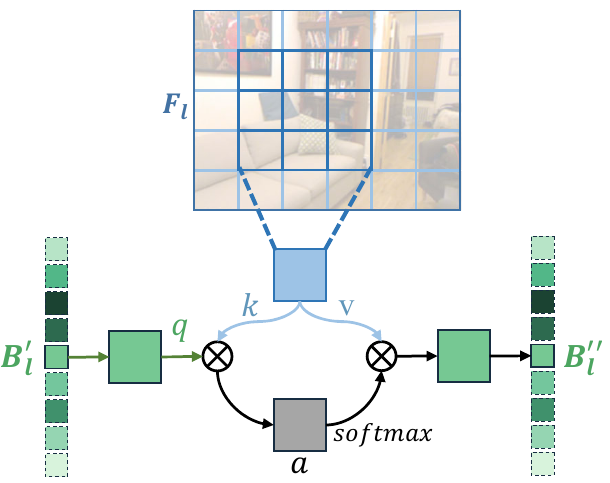}
    \caption{An overview of the cross-attention block, which captures the correlation between features from two branches to refine the bin embedding.
    }
    \label{fig:ca}
\end{figure}

To generate a precise bin partition, global information about the scene is required to represent the depth distribution. However, previous methods~\cite{bhat2021adabins, peng2022pixelwise} only predict a bin partition with high-resolution feature maps in an end-to-end framework which leads to an inevitable loss of global information. To fully utilize the multi-scale depth feature in the latent decoder layer, we introduce an incremental manner to decouple the depth range from coarse to fine in $\boldsymbol{L}$ stages. Firstly, we extract the 3D spatial feature based on the BIM block and adapt the seed bins $\boldsymbol{B}_{seed}$ as the original depth distribution prior for depth decoupling. And as illustrated in Fig.~\ref{fig:trans}, seed bins $\boldsymbol{B}_{seed}$ are sent to a transformer block to interact with the semantic features from the depth modulating branch and aggregate these features from both branches to capture the depth distribution at each scale. Here, we use $\boldsymbol{B}_{l}$ to represent the bin embeddings which are sent to the transformer block at $l~th$ stage, and seed bins $\boldsymbol{B}_{seed}$ are equal to $\boldsymbol{B}_{1}$. Firstly, we leverage the self-attention mechanism to capture the relationship between each spatial geometry feature within the seed bins resulting in a bin feature $\boldsymbol{B}_{l}^{\prime}$ with stronger distribution expression ability:
\begin{equation}\label{self-a}
\boldsymbol{B}_{l}^{\prime}=self\mbox{-}attention(\boldsymbol{B}_{l})=\operatorname{softmax}\left(\frac{\boldsymbol{B}_{l} \times \boldsymbol{B}_{l}^T}{\sqrt{dim}}\right) \boldsymbol{B}_{l}.
\end{equation}
After a layer-normalization operation add with bin embeddings $\boldsymbol{B}_{l}$, we compute the affinity matrix between bin feature $\boldsymbol{B}_{l}^{\prime}$ and the projected depth feature $\boldsymbol{F}_l^{proj} \in h_l*w_l \times dim $ by:
\begin{equation}\label{Q@K}
    \boldsymbol{A} = \frac{\boldsymbol{B}_{l}^{\prime} \times \boldsymbol{F}_l^{proj^T}}{\sqrt{dim}} \in \mathbb{R}^{m_{l}\times h_l*w_l}.
\end{equation}
Each element within the affinity matrix $\boldsymbol{A}$ represents the similarity between depth interval and image patch feature. Then, we normalize $\boldsymbol{A}$ with softmax operation:
\begin{equation}\label{softmax}
    \boldsymbol{A}^{\prime} = \operatorname{softmax}\left(\boldsymbol{A}\right) \in \mathbb{R}^{m\times h_l*w_l}.
\end{equation}
Finally, we generate the bin embedding $\boldsymbol{B}_{l}^{\prime \prime}$ by combining image patch feature related to a specific depth interval based on the affinity matrix $\boldsymbol{A}^{\prime}$:
\begin{equation}\label{softmax}
    \boldsymbol{B}_{l}^{\prime \prime} = \boldsymbol{A}^{\prime} \times \boldsymbol{F}_l^{proj} \in \mathbb{R}^{m_l\times dim}.
\end{equation}
Eq.~\ref{softmax} means that the embedding of a depth interval is represented as a combination of most related image patch features because the characteristic of softmax function amplifies the similarity between depth interval and the most related image patch features. We adapt the cross-attention mechanism to implement the above algorithm as shown in Fig.~\ref{fig:ca}.
And the bin embedding is refined by the feed-forward network:
\begin{equation}
    \hat{\boldsymbol{B}_l} =\max \left(0, \boldsymbol{B}_{l}^{\prime \prime} \boldsymbol{W}_1+b_1\right) \boldsymbol{W}_2+b_2 \in \mathbb{R}^{m_l\times dim},
\end{equation}
which $\boldsymbol{W}_1$, $\boldsymbol{W}_2$ and $b_1$, $b_2$ are the weights and bias of Linear Layers within the feed-forward network.
Subsequently, as shown in the Fig.~\ref{fig:trans}, the refined bin embedding $\hat{\boldsymbol{B}_l}$ is fed into three different paths for guiding probability prediction by inverse projection path; generating depth categories $\textbf{c}(\textbf{b}_{l})$ with a transform layer; splitting $\hat{\boldsymbol{B}_l}$ to $\boldsymbol{B}_{l+1}$ by a Linear Layer for further decoupling. 

\begin{figure}[t]
    \centering
    \includegraphics[width=1\columnwidth]{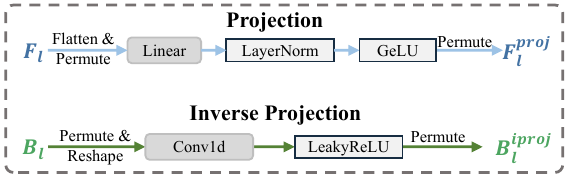}
    \caption{An overview of our projection and inversely projection operations, which project the features from the feature space of the modulating branch to the feature space of the decoupling branch or vice versa.
    }
    \label{fig:proj}
    \vspace{-10pt}
\end{figure}

For the inverse projection, as illustrated in the bottom of Fig.~\ref{fig:proj}, the refined bin embedding~$\hat{\boldsymbol{B}_l}$ is inversely projected to the modulating feature space by a convolution layer and serves as a depth distribution guidance to promote depth modulating. For generating depth categories $c(\textbf{b}_{l})$, we first transform the refined bin embedding $\hat{\boldsymbol{B}_l}$ into bin partition $\textbf{b}_l \in \mathbb{R}^{m_l}$ by $MLPs$:
\begin{equation}\label{bin width}
    \textbf{b}_l = Norm(MLPs(\hat{\boldsymbol{B}_l})).
\end{equation}
The normalized operation is:
\begin{equation}\label{bins partition}
     \textbf{b}_{l}^{\prime}=\frac{ReLU(\textbf{b}_{l})+\epsilon}{\sum(ReLU(\textbf{b}_{l})+\epsilon)},
\end{equation}
where $\epsilon=10^{-3}$ to ensure the width of each bin is strictly positive. 
To represent extreme value which out of the relative depth range of sparse depth map, we add two boundary categories close to the boundary of the dataset's depth range respectively and get the final bin partition $\textbf{b}_l \in \mathbb{R}^{m_{l}+2}$. Then the bin centers $c(\textbf{b}_{l})$ are computed by Eq.~\ref{bins center} as the depth categories. Particularly, we adapt the \textbf{relative depth range} of sparse depth map to further restrict the depth categories instead of the \textbf{absolute depth range}. Because the absolute depth range of the dataset mismatch the depth range of a specific scene, leading to incorrect depth categories generation, as illustrated in Fig.~\ref{fig3}~(a)(c). The depth range of the dataset~(Fig.~\ref{fig3}~(c)) is \textit{0.6-10m} while that of the specific scene is \textit{1.9-3.6m}~(Fig.~\ref{fig3}~(a)). For refining bin embedding $B_{l}$, we increase the length of bin embedding $B_l$ to twice with $MLPs$ followed by the ReLU activation function:
\begin{equation}\label{bins split}
    \boldsymbol{B}_{l+1}=ReLU(MLPs(\boldsymbol{B}_l)).
\end{equation}
And the refined bin embedding $\boldsymbol{B}_{l+1} \in \mathbb{R}^{m_{l+1}}$ ($m_{l+1}=2*m_l$) subsequently sends to the next transformer block to generate a fine-level bin partition incrementally.

\subsection{Adaptive Depth Modulating}
\begin{figure}[b]
    \centering
    \includegraphics[width=1\columnwidth]{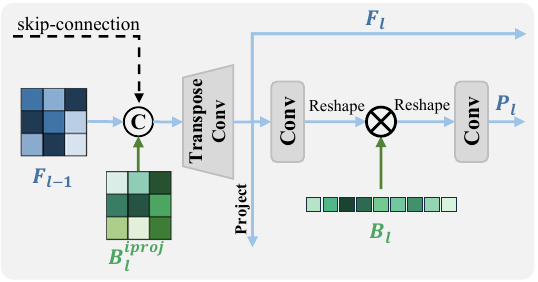}
    \caption{
    An overview of our decoder block architecture, which promotes depth decoupling, refines depth feature and generates probability representations.
    }
    \label{fig:decoder}
\end{figure}

In addition to utilizing the depth feature to guide depth discretizing, we guide the corresponding probability prediction and the depth feature modulation by leveraging the depth distribution information of scenes within bin embedding. Specifically, at each stage $l$, the decoder block $DB_l$ (as illustrated in Fig.~\ref{fig:decoder}) outputs a modulated depth feature $\boldsymbol{F}_l \in \mathbb{R}^{h_l * w_l \times dim}$, receiving the depth feature $\boldsymbol{F}_{l-1}$ from $DB_{l-1}$, the inverse projected bin embedding $\boldsymbol{B}_{l}^{iproj}$ from depth decoupling branch and the skip feature $\boldsymbol{E}_l$ from encoder as input:
\begin{equation}\label{depth modulate}
    \boldsymbol{F}_{l} = DB_l(\boldsymbol{F}_{l-1}, \boldsymbol{B}_{l}^{iproj}, \boldsymbol{E}_l).
\end{equation}

Sequentially, the depth feature $\boldsymbol{F}_{l}$ is projected into bin feature space by a convolution layer to guide the depth decoupling in the next stage, as illustrated in the top of  Fig.~\ref{fig:proj}. 
Meanwhile, guided by the depth distribution information from the bin embedding $\hat{\boldsymbol{B}}_{l}$, we can predict the probabilities of pixels about which depth range they are.
Concretely, we first transform the feature dimension of $\boldsymbol{F}_l$ by a convolution layer and interact it with the $\boldsymbol{B}_{l}$ by matrix multiplication. Then, we generate the probabilistic representation $\boldsymbol{P}_l$ with a convolution layer followed by a soft-max function:
\begin{equation}\label{prob predict}
    \boldsymbol{P}_{l} = softmax(Conv(Conv(\boldsymbol{F}_l)\times \boldsymbol{B}_{l})) \in \mathbb{R}^{h_l \times w_l \times (m_l + 2)} .
\end{equation}

In the final stage, we design a probability prediction block PPB to predict the final per-pixel probabilities, which outputs the finest per-pixel probability map. Similar to the decoder block, it also receives features $\boldsymbol{F}_{l-1}$ from the output of the decoder block in the previous stage, bin embeddings $\boldsymbol{B}_{l}^{iproj}$ reverse-projected from the depth decoupling branch, and skip features $\boldsymbol{E}_l$ passed from the encoder:
\begin{equation}\label{ppb}
    \boldsymbol{P}_{L} = PPB(\boldsymbol{F}_{L-1}, \boldsymbol{B}_{L}^{iproj}, \boldsymbol{E}_L),
\end{equation}
where $\boldsymbol{L}$ is the final progressive stage.
These three types of features are first concatenated together, then passed through a convolutional layer, a batch normalization layer, and a LeakyReLU activation function before interacting with the finest bin embedding $\boldsymbol{B}_{l}$, and generate the final probability map $\boldsymbol{P}_{final}$ in the same manner as Eq.~\ref{prob predict}.

Finally, we can compute each pixel's depth based on the Eq.~\ref{depth predict} and get the depth map $\boldsymbol{D}_l^{pred}$ at the $l$-th depth modulating stage. Additionally, we use cspn postprocessing method~\cite{cheng2018depth} to refine our final depth prediction. Based on this adaptive strategy, we can keep discretizing the depth range into finer categories and predicting a more accurate probability map which leads to a more precise depth map.

\begin{figure}[t]
    \centering
    \includegraphics[width=1\columnwidth]{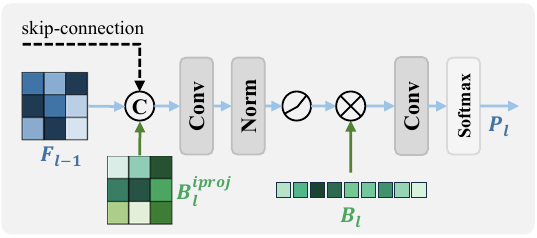}
    \caption{
    An overview of our probabilistic prediction block architecture, which generates final probability representations with full resolution.
    }
    \label{fig:pdb}
    \vspace{-10pt}
\end{figure}

\subsection{Multi-scale Supervision}
\textbf{Pixel-wise depth loss.}
We train the network by using $l_1$ and $l_2$ loss as depth loss as follows:
\begin{equation}\label{depth loss}
    \mathcal{L}_{depth}^{\rho}\left(\mathbf{D}^{gt}, \mathbf{D}^{{pred}}\right)=\frac{1}{|\mathcal{N}|} \sum_{n \in \mathcal{N}}\left|d_n^{gt}-d_n^{{pred}}\right|^\rho,
\end{equation}
where $\mathbf{D}^{gt}$ is the ground truth depth, $\mathbf{D}^{{pred}}$ is the prediction from our algorithm, and $\mathcal{N}$ denote the set of valid pixels in $\mathbf{D}^{gt}$. Here, $\rho$ is set to 1 for $l_1$ loss and 2 for $l_2$ loss.

\noindent \textbf{Bin-center density loss.}
This loss encourages the distribution of bin centers to follow that of depth values in the ground truth. We use the bi-directional Chamerfer Loss~\cite{fan2017point} as a regularizer:
\begin{equation}\label{bin loss}
    \mathcal{L}_{{bins }}=\sum_{d \in \mathbf{D}^{gt}} \min _{c \in c(\mathbf{b})}\|d-c\|^2+\sum_{c \in c(\mathbf{b})} \min _{d \in \mathbf{D}^{gt}}\|d-c\|^2.
\end{equation}
Finally, the loss of multi-scale supervision is:
\begin{equation}
    \mathcal{L}_{MSS} = \sum_{l}^{L} \omega_{l}*(\mathcal{L}_{depth}^{1}+\mathcal{L}_{depth}^{2}+\beta*\mathcal{L}_{bins}).
\end{equation}
where $\omega_{l}$ controls the contribution of different stages, and $\beta$ decides the contribution of $\mathcal{L}_{bins}$. We experimentally set $\omega_{l}= 0.5^{L-l}$ and $\beta= 0.1$ for all our experiments.

\section{Experiments}
\subsection{Datasets} \label{sec:setup}
\textbf{NYU-Depth-v2}~\cite{silberman2012indoor} consists of RGB images, depth maps and surface normal maps collected from 464 indoor scenarios. We follow the standard train-test split, where 249 scenes with 48k training samples are used for training, and the remaining 215 scenes with 654 testing samples for testing. The image of size 640$\times$480 is first down-sampled to half and then center-cropped, producing a 304$\times$228 image. 

\textbf{ScanNet-v2}~\cite{dai2017scannet} is an RGB-D video dataset containing 2.5 million image-depth map pairs, collected from 1513 distinct indoor spaces. We train our network on a subset of around 24K training samples and evaluate the model with 2135 samples. We apply the same data preprocessing and augmentation operations as the NYU-Depth-v2 dataset.

\textbf{KITTI Depth Completion Dataset}~\cite{uhrig2017sparsity} is a large outdoor dataset. It consists of 86K RGB images and LiDAR pairs for training, 1K pairs for validation and 1K pairs for testing. Same as~\cite{kam2022costdcnet}, we center-cropped the images with size of $1216 \times 240$ to ignore the regions without LiDAR measurements.

\begin{figure*}[t]
    \centering
    \includegraphics[width=1\linewidth]{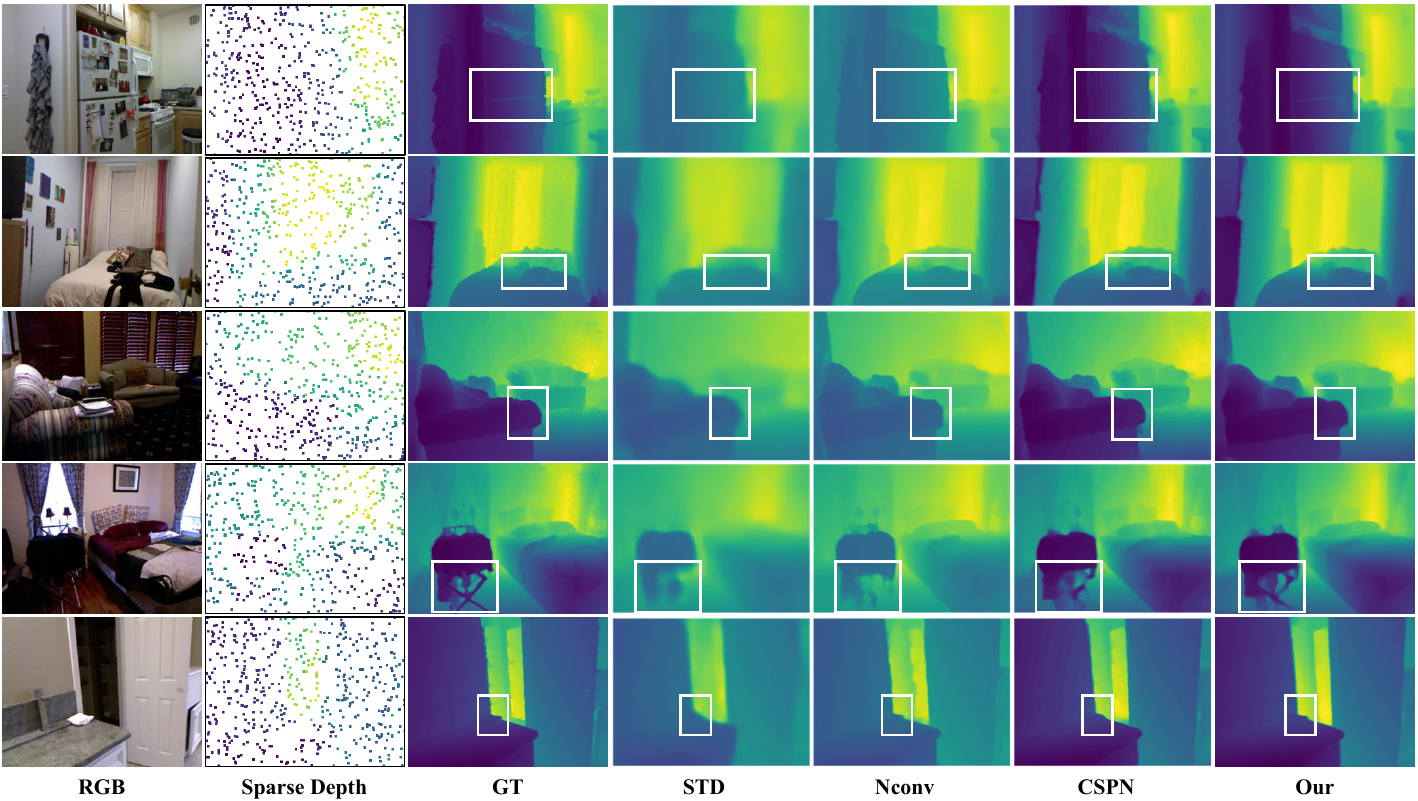}
    \caption{Qualitative results on NYU-Depth-v2 dataset. The images from left to right are RGB image, the sparse depth, the ground truth, result from~\cite{ma2018sparse}, result from~\cite{eldesokey2019confidence}, result from~\cite{kam2022costdcnet}, result from~\cite{conti2023sparsity}, result of our full model.}
    \label{fig:nyu}
    \vspace{-10pt}
\end{figure*}

\subsection{Implementation details}
We implement experiments on PyTorch and train with RTX 3090Ti GPU with the Adam optimizer for 40 epochs. We use a batch size of 8 and 1-cycle policy schedule for adjusting the learning rate with a max learning rate of 0.00357. Following previous works~\cite{cheng2018depth}, we randomly sample 500 reference depth samples from the sparse depth map, and evaluate our method with 6 metrics: relative mean absolute error (AbsRel); Root Mean Squared Error (RMSE); average ($log_{10}$) error; log Root Mean Squared Error (RMSE log); Squared Relative Difference (Sq. Rel) and threshold accuracy $\delta_i$ and 2 additional metrics Inverse RMSE (IRMSE); Inverse MAE (IMAE) for KITTI dataset.

For reproducing STD~\cite{ma2018sparse}, Nconv~\cite{eldesokey2019confidence}, CSPN~\cite{cheng2018depth}, NeWCRF~\cite{yuan2022neural}, P3Depth~\cite{patil2022p3depth}, Sparsity~\cite{conti2023sparsity}, CostDCNet~\cite{kam2022costdcnet} and Adabins~\cite{bhat2021adabins}, we use the source codes or the pretrained model released online from the author. 
And the results of other methods~\cite{qiu2019deeplidar, liu2021learning, liu2021fcfr, zhao2021adaptive, tang2020learning, imran2019depth, lee2021depth, tran2023adaptive, basak2020monocular, nazir2022semattnet, ma2019self} are copied from the original paper.

\renewcommand\arraystretch{1.2}
\begin{table}[b]
\small
\setlength{\abovecaptionskip}{0.01cm}
\setlength{\belowcaptionskip}{0.01cm}
\centering
\caption{The result of depth completion on {NYU-Depth-v2}.
The best results are in bold, and the second best results are underlined.}
\resizebox{\columnwidth}{!}{
\begin{tabular}{c| c c | c c c}
\toprule[2pt]
\multicolumn{1}{c|}{\multirow{2}{*}{\textbf{Methods}}} & \multicolumn{2}{c|}{\textbf{Lower is better}} & \multicolumn{3}{c}{\textbf{Higher is better}} \\
\cline{2-6}
\multicolumn{1}{c|}{} 
& {AbsRel $\downarrow$} 
& {RMSE $\downarrow$} 
& $\delta_{1.25}$ $\uparrow$
& $\delta_{1.25^2}$ $\uparrow$
& $\delta_{1.25^3}$ $\uparrow$ \\
\hline
STD~\cite{ma2018sparse} & {0.053} & {0.240} & {0.981} & {0.996} & \textbf{0.999} \\ 
Nconv~\cite{eldesokey2019confidence} & {0.017} & {0.123} & {0.991} & \underline{0.998} & \textbf{0.999} \\
CSPN~\cite{cheng2018depth} & {0.016} & {0.117} & {0.992} & \textbf{0.999} & \textbf{0.999} \\
DeepLiDAR~\cite{qiu2019deeplidar} & {0.022} & {0.115} & {0.993} & \textbf{0.999}  & \textbf{0.999} \\
S-KNet~\cite{liu2021learning} & {0.015} & {0.111} & {0.993} & \textbf{0.999} & \textbf{0.999} \\
FCFR-Net~\cite{liu2021fcfr} & {0.015} & {0.106} & \textbf{0.995} & \textbf{0.999} & \textbf{0.999} \\
ACMNet~\cite{zhao2021adaptive} & {0.015} & {0.105} & \underline{0.994} & \textbf{0.999} & \textbf{0.999} \\
GuideNet~\cite{tang2020learning} & {0.015} & {0.101} & \textbf{0.995} & \textbf{0.999} & \textbf{0.999} \\
NeWCRF~\cite{yuan2022neural} & {0.034} & {0.217} & {0.976} & \textbf{0.996} & \textbf{0.999} \\
P3Depth~\cite{patil2022p3depth} & {0.025} & {0.141} & {0.990} & \textbf{0.999} & \textbf{0.999} \\
Sparsity~\cite{conti2023sparsity} & {0.015} & {0.138} & {0.993} & \textbf{0.999}  & \textbf{0.999}\\
\hline
DC-all~\cite{imran2019depth} &  \textbf{0.013} & {0.118} & {0.978} & {0.994}  & \textbf{0.999} \\
Adabins~\cite{bhat2021adabins} & {0.018} & {0.117} & {0.993} & \textbf{0.999}  & \textbf{0.999}\\
Lee \textit{et.al}~\cite{lee2021depth} & \underline{0.014} & {0.104} & \underline{0.994} & \textbf{0.999} & \textbf{0.999}\\
CostDCNet~\cite{kam2022costdcnet} & \textbf{0.013} & \underline{0.097} & \textbf{0.995} & \textbf{0.999} & \textbf{0.999} \\
\textbf{Ours} & \underline{0.014} & \textbf{0.095} & \textbf{0.995} & \textbf{0.999} & \textbf{0.999}\\
\bottomrule[2pt]
\end{tabular}
}
\label{tab:nyu}
\end{table}

\subsection{Quantitative Results} \label{sec:result_nyu}
\subsubsection{Results on NYU-Depth-v2} 
The quantitative results on NYU-Depth-v2 are presented in Table~\ref{tab:nyu}. The methods are categorized into regression-based (top) and classification-based (bottom) approaches. Our method demonstrates competitive performance across both categories. Compared to DeepLiDAR \cite{qiu2019deeplidar} and P3Depth \cite{patil2022p3depth}, our method surpasses them on all metrics without the help of additional geometry representation constraints they leverage which shows the effectiveness of our progressive depth decoupling and modulating strategy. Compared to classification-based methods like CostDCNet \cite{kam2022costdcnet}, our method achieves higher precision in depth prediction as indicated by $RMSE$ and $\delta_{1.25}$. Our lightweight BIM captures 3D scene information effectively, promotes the bin partition and involves more depth distribution priors. And our progressive approach facilitates interaction between depth decoupling and modulating branches, gradually producing a fine-level depth map.

\subsubsection{Results on KITTI} We evaluate our method on KITTI dataset and compare it with other methods, as shown in Table~\ref{tab:kitti}. The results indicate that the performance of our method is outstanding to other methods. This indicates that our method is also applicable to more challenging outdoor scenarios. Based on the progressive depth decoupling strategy, our method can also accurately predict depth categories in outdoor scenes, ultimately achieving commendable performance.

\renewcommand\arraystretch{1.2}
\begin{table}[b]
\small
\setlength{\abovecaptionskip}{0.01cm}
\setlength{\belowcaptionskip}{0.01cm}
\centering
\caption{The result of depth completion on {KITTI}.
The best results are in bold, and the second best results are underlined.}
\resizebox{\columnwidth}{!}{
\begin{tabular}{c| c c c c}
\toprule[2pt]
\textbf{Methods}
& {MAE $\downarrow$} 
& {RMSE $\downarrow$} 
& {iRMSE $\downarrow$}
& {iMAE $\downarrow$} \\
\hline
SemAttNet(bb)~\cite{nazir2022semattnet} & 796.26 & 1344.37 & 7.08 & 4.96 \\
SemAttNet~\cite{nazir2022semattnet} & 796.26 & 1344.37 & 7.08 & 4.96 \\
Sparse-to-Dense(gd)~\cite{ma2019self} & 524.35 & 1298.24 & 6.84 & 3.19 \\
Nconv~\cite{eldesokey2019confidence} & \underline{360.28} & {1268.22} & \underline{4.67} & \underline{1.52} \\
Sparse-to-Dense~\cite{ma2019self} & 501.21 & \underline{1247.48} & 6.14 & 2.91 \\
\textbf{Ours} & \textbf{294.63} & \textbf{1107.29} & \textbf{3.65} & \textbf{1.31}\\
\bottomrule[2pt]
\end{tabular}
}
\label{tab:kitti}
\end{table} 

\subsubsection{Results on  ScanNet-v2} The comparison results between our method and SOTA methods are illustrated in Table~\ref{tab:scannet}. In general, our method exceeds them in all metrics. Specifically, compared with P3Depth~\cite{patil2022p3depth}, our method improves the $RMSE$ by $11.69 \%$ and the $\delta_{1.25}$ by $2.6 \%$. 
These results on this dataset with more diverse scenarios also demonstrate the effectiveness of our method.

\renewcommand\arraystretch{1.2}
\begin{table}[b]
\small
\setlength{\abovecaptionskip}{0.01cm}
\setlength{\belowcaptionskip}{0.01cm}
\centering
\caption{The quantitative results on {ScanNetv2}. The best results are in bold, and the second best results are underlined.}
\resizebox{\columnwidth}{!}{
\begin{tabular}{c| c c | c c c}
\toprule[2pt]
\multicolumn{1}{c|}{\multirow{2}{*}{\textbf{Methods}}} & \multicolumn{2}{c|}{\textbf{Lower is better}} & \multicolumn{3}{c}{\textbf{Higher is better}} \\
\cline{2-6}
& {AbsRel $\downarrow$} 
& {RMSE $\downarrow$} 
& {$\delta_{1.25}$ $\uparrow$} 
& {$\delta_{1.25^2}$ $\uparrow$}  
& {$\delta_{1.25^3}$ $\uparrow$} \\
\hline
STD~\cite{ma2018sparse} & {0.394} & {0.302} & {0.929} & {0.959} & {0.970}\\  
Nconv~\cite{eldesokey2019confidence} & {0.361} & {0.286} & \underline{0.939} & {0.959} & {0.969}\\
CSPN~\cite{cheng2018depth} & {0.348} & \underline{0.276} & \textbf{0.946} & \underline{0.961} & {0.970}\\
P3Depth~\cite{patil2022p3depth} & {0.320} & {0.277} & {0.921} & {0.956} & \underline{0.971} \\
\textbf{Ours} & \textbf{0.305} & \textbf{0.248} & \textbf{0.946} & \textbf{0.963} & \textbf{0.973}\\
\bottomrule[2pt]
\end{tabular}
}
\label{tab:scannet}
\end{table}

\subsubsection{Muti-scale Depth Predictions}
The results in Table~\ref{tab:multi} show the depth prediction performance at each stage which reflects our progressive depth completion framework can gradually refine the depth prediction. Meanwhile, these results also show that our method is suitable for the flexibility of depth map quality requirements.
\begin{figure*}[t]
    \centering
    \includegraphics[width=1\linewidth]{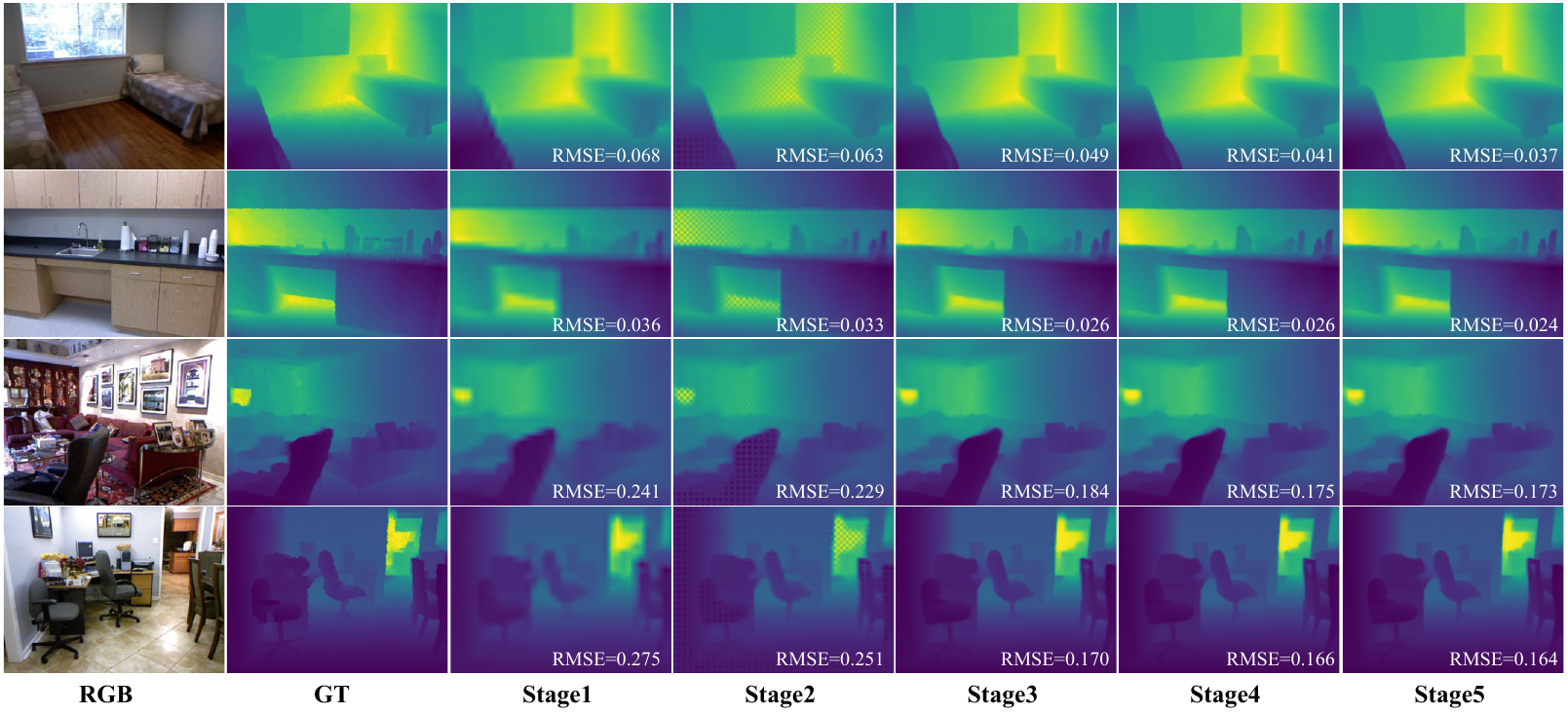}
    \caption{Our multi-stages depth predictions results on NYU-Depth-v2 dataset. The images from left to right are RGB image and the depth prediction of each progressive stage.}
    \label{fig:multi_depth}
    \vspace{-10pt}
\end{figure*}

\renewcommand\arraystretch{1.2}
\begin{table}[b]
\small
\setlength{\abovecaptionskip}{0.01cm}
\setlength{\belowcaptionskip}{0.01cm}
\centering
\caption{The ablation studies result of multi-stages prediction on NYU-Depth-v2.}
\resizebox{\columnwidth}{!}{
\begin{tabular}{c|cc|ccc}
\toprule[2pt]
\multicolumn{1}{c|}{\multirow{2}{*}{\textbf{Methods}}} & \multicolumn{2}{c|}{\textbf{Lower is better}} & \multicolumn{3}{c}{\textbf{Higher is better}} \\
\cline{2-6}         
& AbsRel $\downarrow$
& RMSE $\downarrow$
& $\delta_{1.25}$ $\uparrow$
& $\delta_{1.25^2}$ $\uparrow$
& $\delta_{1.25^3}$ $\uparrow$ \\
\hline
Stage1 & 0.027 & 0.135 & \underline{0.992} & \textbf{0.999} & \textbf{0.999} \\
Stage2 & 0.032 & 0.142 & {0.982} & \underline{0.998} & \textbf{0.999} \\
Stage3 & 0.017 & 0.102 & \textbf{0.995} & \textbf{0.999} & \textbf{0.999}\\
Stage4 & \underline{0.015} & \underline{0.097} & \textbf{0.995} & \textbf{0.999} & \textbf{0.999} \\
Stage5 & \textbf{0.014} & \textbf{0.095} & \textbf{0.995} & \textbf{0.999} & \textbf{0.999} \\
\bottomrule[2pt]
\end{tabular}
}
\label{tab:multi}
\end{table}

\subsubsection{Robustness to Depth Sample Numbers}
\begin{figure}[h]
    \centering
    \includegraphics[width=0.8\columnwidth]{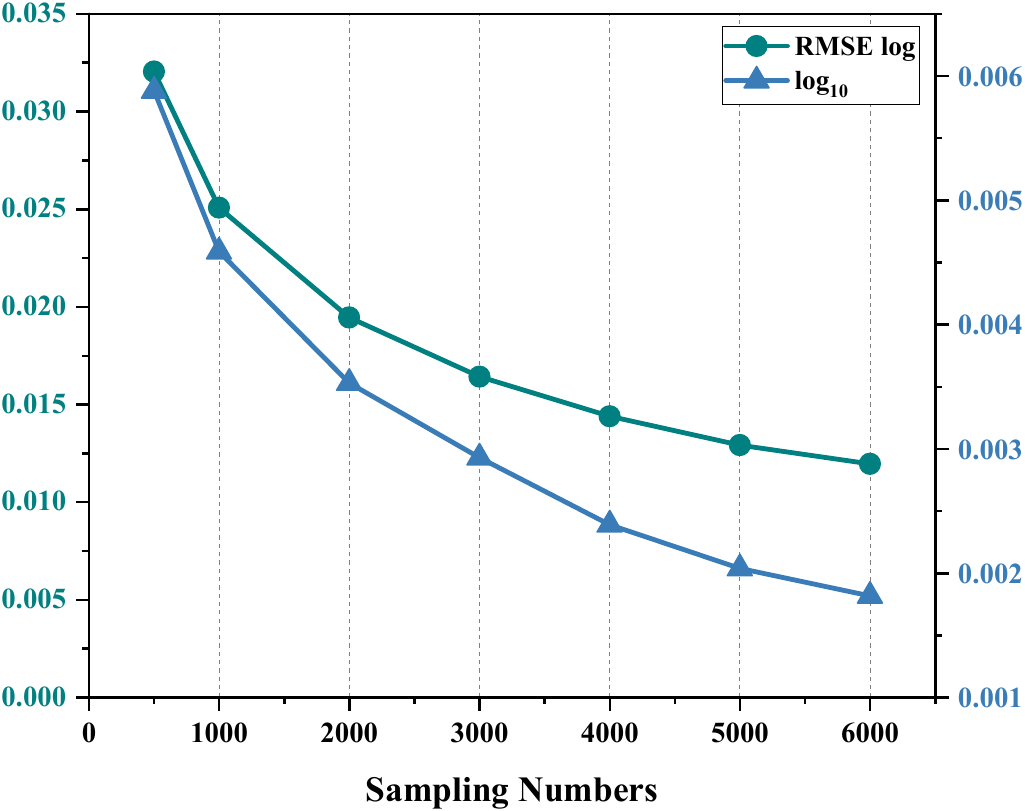}
    \caption{Generalization with respect to sampling numbers on NYU-Depth-v2. The performance is measured by $RMSE log$ and $log_{10}$.}
    \label{fig:samples_tim}
    \vspace{-10pt}
\end{figure}

To verify the robustness of our method for different numbers of sampling points, we conduct experiments with different sampling numbers from 500 to 6000. As shown in Fig.~\ref{fig:samples_tim}, the performance represented by typical metrics $RMSE$ and $\delta_{1.25}$ of our method first rises rapidly and then gradually converges. All metrics results are shown in Table~\ref{tab:samples}. Furthermore, we compare our methods with different depth completion methods under 6000 points in Table~\ref{tab:nyu6000}. We also far outperform other methods in $MAE$ and $RMSE$ metrics, which indicates that our method is more accurate about overall depth prediction.

\renewcommand\arraystretch{1.2}
\begin{table}[b]
\small
\setlength{\abovecaptionskip}{0.01cm}
\setlength{\belowcaptionskip}{0.01cm}
\centering
\caption{The sample number ablation studies result of depth prediction on NYU-Depth-v2.}
\resizebox{\columnwidth}{!}{
\begin{tabular}{c|cc|ccc}
\toprule[2pt]
\multicolumn{1}{c|}{\multirow{2}{*}{\textbf{Samples}}} & \multicolumn{2}{c|}{\textbf{Lower is better}} & \multicolumn{3}{c}{\textbf{Higher is better}} \\
\cline{2-6}         
& AbsRel $\downarrow$
& RMSE $\downarrow$
& $\delta_{1.25}$ $\uparrow$
& $\delta_{1.25^2}$ $\uparrow$
& $\delta_{1.25^3}$ $\uparrow$ \\
\hline
500 & 0.014 & 0.095 & {0.995} & \textbf{0.999} & \textbf{0.999} \\
1000 & 0.011 & 0.075 & \underline{0.997} & \textbf{0.999} & \textbf{0.999} \\
2000 & 0.008 & 0.059 & \textbf{0.999} & \textbf{0.999} & \textbf{0.999}\\
3000 & {0.007} & {0.050} & \textbf{0.999} & \textbf{0.999} & \textbf{0.999} \\
4000 & {0.006} & {0.044} & \textbf{0.999} & \textbf{0.999} & \textbf{0.999} \\
5000 & \underline{0.005} & \underline{0.039} & \textbf{0.999} & \textbf{0.999} & \textbf{0.999} \\
6000 & \textbf{0.004} & \textbf{0.036} & \textbf{0.999} & \textbf{0.999} & \textbf{0.999} \\
\bottomrule[2pt]
\end{tabular}
}
\label{tab:samples}
\end{table}

\renewcommand\arraystretch{1.2}
\begin{table}[b]
\small
\setlength{\abovecaptionskip}{0.01cm}
\setlength{\belowcaptionskip}{0.01cm}
\centering
\caption{The quantitative results on {NYU-DEPTH-v2} under 6000 samples. The best results are in bold, and the second best results are underlined.}
\resizebox{\columnwidth}{!}{
\begin{tabular}{c|c|c c}
\toprule[2pt]
\textbf{Methods} & \textbf{Sampling Numbers} & {MAE $\downarrow$} & {RMSE $\downarrow$}\\
\hline
STD~\cite{ma2018sparse} & 6000 & {0.087} & {0.185} \\
Monocular~\cite{basak2020monocular} & 6000 & {0.103} & {0.388}\\
CSPN~\cite{cheng2018depth} & 6000 & {0.022} & {0.07} \\
AAF~\cite{tran2023adaptive} & 6000 & \underline{0.019} & \underline{0.055} \\
\textbf{Ours} & 6000 & \textbf{0.015} & \textbf{0.036} \\
\bottomrule[2pt]
\end{tabular}
}
\label{tab:nyu6000}
\end{table}

\subsection{Ablation Studies}

\renewcommand\arraystretch{1.2}
\begin{table}[bh]
\small
\setlength{\abovecaptionskip}{0.01cm}
\setlength{\belowcaptionskip}{0.01cm}
\centering
\caption{The comparison of different depth discretizing methods on {NYU-Depth-v2}. \textit{UD} and \textit{SID} are hand-crafted discretization strategies, \textit{Single stage} is a learning based method with auxiliary network. The bottom part is our method and some variants.}
\resizebox{\columnwidth}{!}{
\begin{tabular}{c|cc|ccc}
\toprule[2pt]
\multicolumn{1}{c|}{\multirow{2}{*}{\textbf{Methods}}} & \multicolumn{2}{c|}{\textbf{Lower is better}} & \multicolumn{3}{c}{\textbf{Higher is better}} \\
\cline{2-6}        
& AbsRel $\downarrow$
& RMSE $\downarrow$
& $\delta_{1.02}$ $\uparrow$
& $\delta_{1.05}$ $\uparrow$
& $\delta_{1.10}$ $\uparrow$ \\
\hline
UD          & 0.016             & 0.106             & {0.819} & {0.939} & {0.976} \\
SID         & \underline{0.015} & 0.105             & {0.840} & {0.943} & \underline{0.977} \\
\hline
Single stage & \underline{0.015} & {0.101} & {0.835} & {0.946} & \textbf{0.978} \\
\hline
Bins\_abs   & \underline{0.015} & 0.099             & {0.844} & {0.945} & \underline{0.977} \\
Bins\_fixed & \textbf{0.014}    & \underline{0.096} & \underline{0.854} & \underline{0.947} & \textbf{0.978} \\
\textbf{Ours}        & \textbf{0.014}    & \textbf{0.095}    & \textbf{0.857} & \textbf{0.948} & \textbf{0.978} \\
\bottomrule[2pt]
\end{tabular}
}
\label{tab:pdd}
\end{table}

\subsubsection{Depth discretiztion methods}
We first examine our incremental depth decoupling branch by replacing it with hand-crafted depth discretization strategies: Uniform Discretization ``UD“ and Spacing-Increasing Discretization ``SID"~\cite{fu2018deep}. 
We also implement the learning based method ``Single stage" as a comparison, which conducts bins partition with a highest resolution feature map from the end of network by an auxiliary network. 
As shown in Table~\ref{tab:pdd}, our method outperforms them on $AbsRel$ and $RMSE$ with a large margin, benefiting from the capability of representing depth distribution by adaptive depth discretizing.
We further compare our relative depth range-constrained depth discretizing method ``Ours" with the absolute depth range-constrained approach ``Bins\_abs". As we can see, performance significantly improves when constraining the depth discretization range according to the sparse depth map. 
In ``Bins\_fixed", we fix the number of bins $m_l$ at each stage to $m_5$. The degraded performance endorses the superiority of our incremental depth decoupling method over direct bin partitioning.

\subsubsection{Bins initializing methods}
We evaluate the effectiveness of BIM by removing it from the depth decoupling branch and randomly initialize seed bins (``Random” in Fig.~\ref{fig:bim_params}). Results demonstrate that BIM generates more precise initial seed bins by extracting the position distribution information from the sparse depth map.
And subsequent decoupling processes are more efficient with the depth prior in it. 
Furthermore, we also compare BIM with traditional convolutional neural networks including ResNet18 and ResNet34 as illustrated in Fig.~\ref{fig:bim_params}. Compared with ResNet34, our BIM shows competitive performance with only 58\% parameters which endorses its effectiveness.

\begin{figure*}[th]
    \centering
    \includegraphics[scale=0.65]{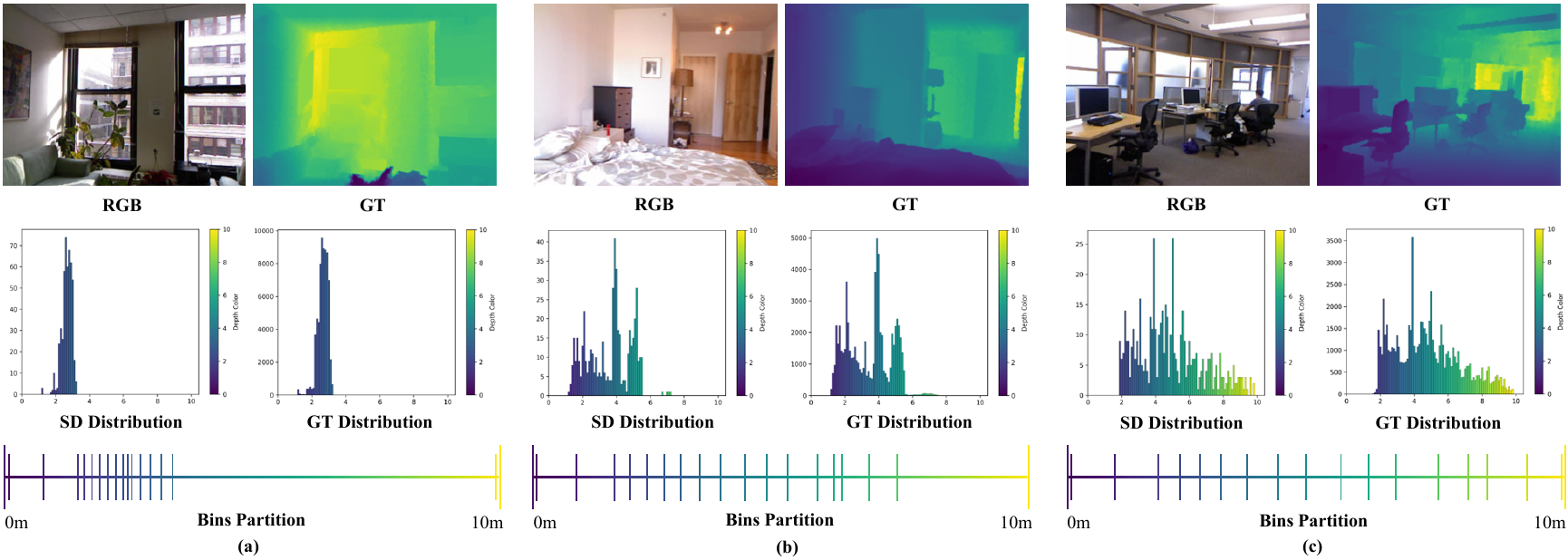}
    \caption{ 
    The illustration of our bin partitions and scene depth distributions in three typical scenes with different depth scopes: (a) narrow, (b) middle, (c) wide. The predicted bin partition on different scenes is close to the corresponding depth distribution.
    }
    \label{fig:bins}
    \vspace{-10pt}
\end{figure*}

\begin{figure}[t]
    \centering
    \includegraphics[width=0.8\columnwidth]{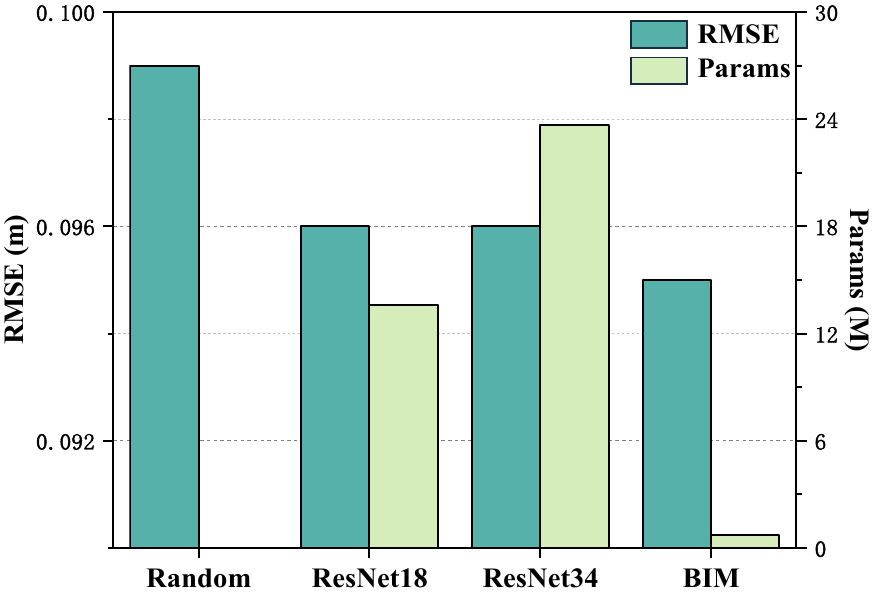}
    \caption{ 
    The comparison of RMSE and Parameters for different bins initializing methods on NYU-Depth-v2.
    }
    \label{fig:bim_params}
    \vspace{-10pt}
\end{figure}

\renewcommand\arraystretch{1.2}
\begin{table}[b]
\small
\setlength{\abovecaptionskip}{0.01cm}
\setlength{\belowcaptionskip}{0.01cm}
\centering
\caption{The ablation studies result of depth completion on {NYU-Depth-v2}.}
\resizebox{\columnwidth}{!}{
\begin{tabular}{c|cc|ccc}
\toprule[2pt]
\multicolumn{1}{c|}{\multirow{2}{*}{\textbf{Methods}}} & \multicolumn{2}{c|}{\textbf{Lower is better}} & \multicolumn{3}{c}{\textbf{Higher is better}} \\
\cline{2-6}         
& AbsRel $\downarrow$
& RMSE $\downarrow$
& $\delta_{1.02}$ $\uparrow$
& $\delta_{1.05}$ $\uparrow$  
& $\delta_{1.10}$ $\uparrow$ \\
\hline
w/o incremental & \underline{0.015} & {0.101} & {0.835} & {0.946} & \textbf{0.978} \\
D2M only & {0.016} & {0.100} & {0.840} & {0.946} & \underline{0.977} \\
M2D only & \underline{0.015} & {0.100} & {0.826} & {0.944} & \textbf{0.978}  \\
w/o $\mathcal{L}_{MSS}$ & \textbf{0.014} & \underline{0.097} & \underline{0.854} & \textbf{0.948} & \textbf{0.978} \\
w/o PPB & \underline{0.015} & {0.099} & {0.849} & \underline{0.947} & \textbf{0.978}\\
\textbf{Ours}  & \textbf{0.014} & \textbf{0.095} & \textbf{0.857} & \textbf{0.948} & \textbf{0.978} \\
\bottomrule[2pt]
\end{tabular}
}
\label{tab:ablation}
\vspace{-10pt}
\end{table}

\renewcommand\arraystretch{1.2}
\begin{table}[b]
\small
\setlength{\abovecaptionskip}{0.01cm}
\setlength{\belowcaptionskip}{0.01cm}
\centering
\caption{The ablation studies result of different skip-connection strategies on {NYU-Depth-v2}.}
\resizebox{\columnwidth}{!}{
\begin{tabular}{c|cc|ccc}
\toprule[2pt]
\multicolumn{1}{c|}{\multirow{2}{*}{\textbf{Methods}}} & \multicolumn{2}{c|}{\textbf{Lower is better}} & \multicolumn{3}{c}{\textbf{Higher is better}} \\
\cline{2-6}         
& AbsRel $\downarrow$
& RMSE $\downarrow$
& $\delta_{1.02}$ $\uparrow$
& $\delta_{1.05}$ $\uparrow$  
& $\delta_{1.10}$ $\uparrow$ \\
\hline
w/o 1st & \textbf{0.014} & {0.097} & {0.853} & \textbf{0.948} & \textbf{0.978} \\
w/o 2nd & \textbf{0.014} & {0.097} & {0.854} & \textbf{0.948} & \textbf{0.978}  \\
w/o 3rd & \textbf{0.014} & \underline{0.096} & \underline{0.855} & \textbf{0.948} & \textbf{0.978} \\
w/o 4th & \textbf{0.014} & {0.097} & \textbf{0.857} & \textbf{0.948} & \textbf{0.978} \\
w/o 1st\&2nd  & \textbf{0.014} & {0.098} & {0.850} & \textbf{0.948} & \textbf{0.978} \\
w/o 3rd\&4th  & \textbf{0.014} & {0.098} & {0.849} & \underline{0.946} & \textbf{0.978} \\
w/o all  & \underline{0.015} & {0.100} & {0.827} & {0.943} & \underline{0.977} \\
w/ all  & \textbf{0.014} & \textbf{0.095} & \textbf{0.857} & \textbf{0.948} & \textbf{0.978} \\
\bottomrule[2pt]
\end{tabular}
}
\label{tab:skip}
\vspace{-10pt}
\end{table}

\subsubsection{Progressive Strategies}
Then we study the effectiveness of our progressive strategy by using the initial bin partition from BIM as the final depth discretization result (``w/o incremental" in Table~\ref{tab:ablation}).
Clearly, the performance decreases dramatically when refining bin partition without the multi-scale feature from global to local. We further investigate the effect of progressive guidance between depth decoupling and modulating branches.
By removing the guidance from depth modulating branch to decoupling branch ``D2M only" and vice versa ``M2D only" respectively, we find that no matter which branch is removed, the performance is significantly decreased. Thus the progressive guidance between both branches is essential to classification-based depth completion methods.

\subsubsection{Multi-scale Supervision} 
As shown in Table~\ref{tab:ablation}, compared with single-scale supervision at the final stage ``w/o $\mathcal{L}_{MSS}$", our method improves the performance by additional intermediate supervision to the multi-scale depth maps.

\subsubsection{Number of Bins}
\begin{figure}[h]
    \centering
    \includegraphics[width=\columnwidth]{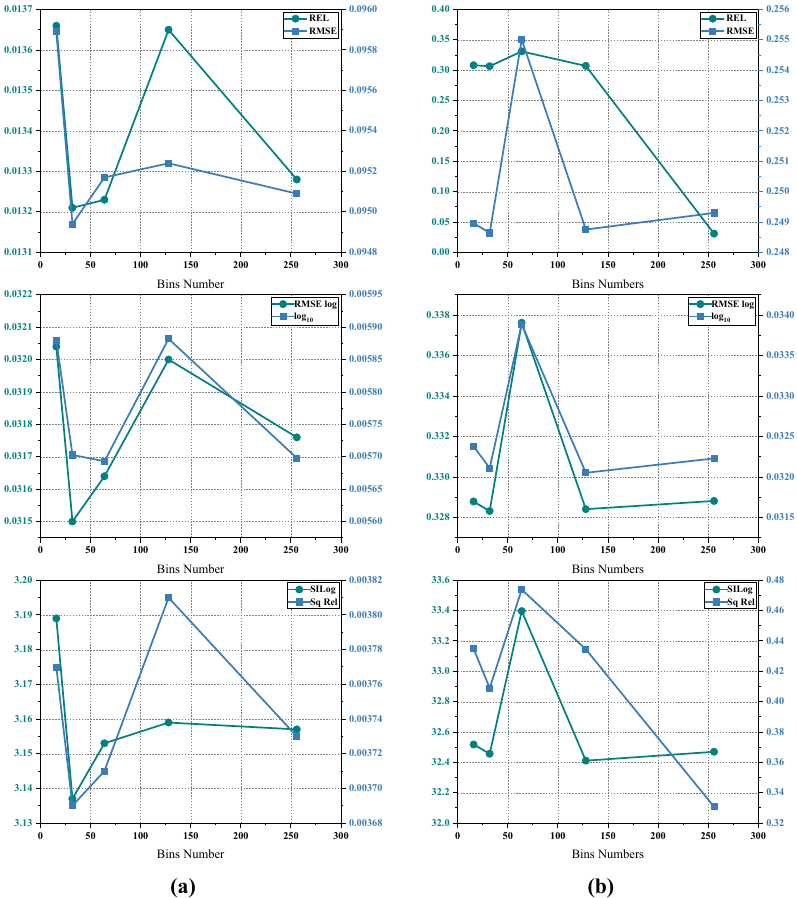}
    \caption{Effect of the number of bins (N) on performance as measured on NYU-Depth-v2 (a) and ScanNet-v2 (b).
    }
    \label{fig:bins_num}
    \vspace{-10pt}
\end{figure}

To study the influence of the number of bins, we train our network on both datasets with various numbers $N$ and measure the performance in terms of $RMSE$. Results are plotted in Fig.~\ref{fig:bins_num}. Interestingly, the relationship between performance and quantity is not strictly linear. Beginning at N = 16, the error decreases initially, followed by a notable increase. As we keep increasing N to 256, the performance variation becomes smaller. These results indicate that around 32 bins are enough to represent the whole scene depth distribution of these two datasets. We select N = 16 for the purpose of effectiveness.

\subsection{Number of Progressive Stages}
We design a progressive depth decoupling and modulating framework to induce the difficulty of directly generating precise depth categories. To explore the effect of the number of progressive stages, we train our network with different progressive stages. As shown in Table~\ref{tab:stages}, the performance of the model is observed to undergo a rapid enhancement as it transitions from one progressive stage to two progressive stages, followed by a gradual convergence. Meanwhile, apart from the initialization stage of depth categories (referred to as bins), the number of depth categories can represent doubles after each stage, thereby realizing our principle of progressive depth decoupling from easy to difficult. However, with the increase of stages, there is also a corresponding increase in the model parameter count. Therefore, for the sake of balancing performance and parameter count, we eventually perform for five stages.

\subsection{Skip-connection Stratigies}
Skip-connection is a commonly employed technique within encoder-decoder structures. In order to assess the impact of it on our network, we conducted ablation experiments on various skip-connection strategies. The experimental results are presented in Table~\ref{tab:skip}. We follow the order of skip connections as shown in Fig.~\ref{fig2} for result demonstration. The first four rows of experiments represent the results of removing individual skip connection routes, indicating that the first and third skip connections have the most significant impact on performance. Subsequently, we conducted experiments removing the first two skip connections routes~(representing shallow features) and the latter two routes~(representing deep features), respectively. The results indicate that deep features contribute more significantly to the performance. Finally, by removing all routes, there was a significant decline in the model's performance. Skip connections, as a simple yet effective method widely used in encoder-decoder structured networks, have been validated for their effectiveness within our framework through these ablation experiments.

\renewcommand\arraystretch{1.2}
\begin{table}[b]
\small
\setlength{\abovecaptionskip}{0.01cm}
\setlength{\belowcaptionskip}{0.01cm}
\centering
\caption{The ablation studies result of the numbers of different progressive stages on NYU-Depth-v2.}
\resizebox{\columnwidth}{!}{
\begin{tabular}{c|c|cc|ccc}
\toprule[2pt]
\multicolumn{1}{c|}{\multirow{2}{*}{\textbf{Methods}}} & \multicolumn{1}{c|}{\multirow{2}{*}{\textbf{Params}}} & \multicolumn{2}{c|}{\textbf{Lower is better}} & \multicolumn{3}{c}{\textbf{Higher is better}} \\
\cline{3-7}  
&
& AbsRel $\downarrow$
& RMSE $\downarrow$
& $\delta_{1.25}$ $\uparrow$
& $\delta_{1.25^2}$ $\uparrow$
& $\delta_{1.25^3}$ $\uparrow$ \\
\hline
Stage1 & 26.2M & 0.033 & 0.166 & \underline{0.984} & \underline{0.997} & \textbf{0.999} \\
Stage2 & 27.9M& \underline{0.015} & 0.100 & \textbf{0.995} & \textbf{0.999} & \textbf{0.999} \\
Stage3 & 29.1M& \underline{0.015} & 0.100 & \textbf{0.995} & \textbf{0.999} & \textbf{0.999}\\
Stage4 & 30.4M& \underline{0.015} & {0.099} & \textbf{0.995} & \textbf{0.999} & \textbf{0.999} \\
Stage5 & 32.4M& \textbf{0.013} & {0.096} & \textbf{0.995} & \textbf{0.999} & \textbf{0.999} \\
Stage6 & 34.4M& \textbf{0.013} & \underline{0.095} & \textbf{0.995} & \textbf{0.999} & \textbf{0.999}\\
Stage7 & 36.4M& \textbf{0.013} & \underline{0.095} & \textbf{0.995} & \textbf{0.999} & \textbf{0.999}\\
Stage8 & 38.5M& \textbf{0.013} & \textbf{0.094} & \textbf{0.995} & \textbf{0.999} & \textbf{0.999}\\
\bottomrule[2pt]
\end{tabular}
}
\label{tab:stages}
\vspace{-10pt}
\end{table}

\begin{figure}[t]
    \centering
    \includegraphics[width=\columnwidth]{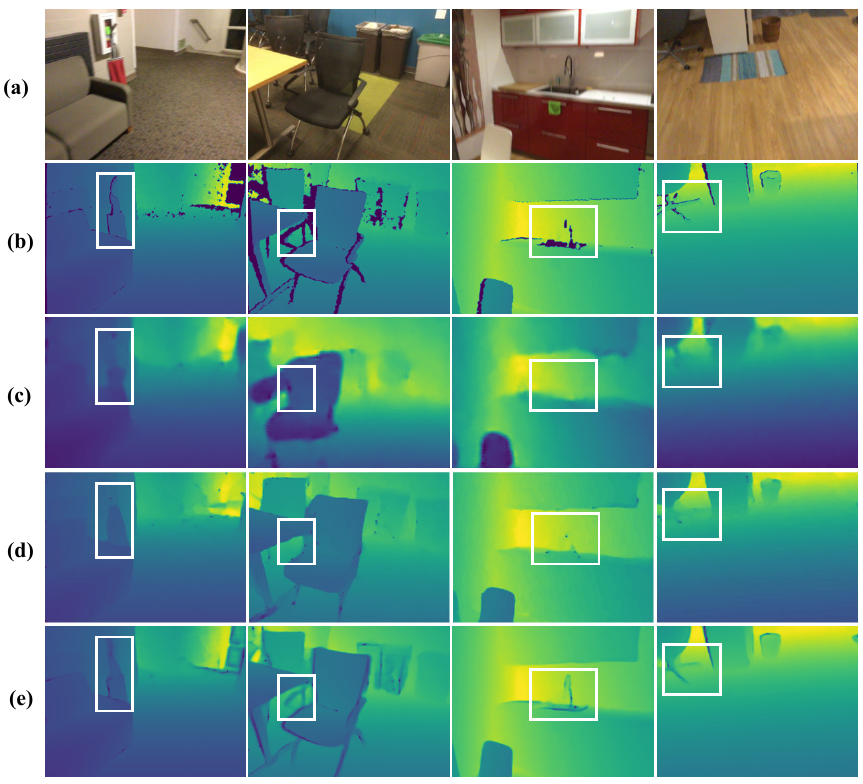}
    \caption{
    Qualitative results on ScanNet-v2 dataset. From top to down is (a) RGB image, (b) the ground truth, (c) result from~\cite{eldesokey2019confidence}, (d) result from~\cite{cheng2018depth}, (e) result of our method.
    }
    \label{fig:scan}
    \vspace{-10pt}
\end{figure}

\begin{figure*}[!t]
    \centering
    \includegraphics[scale=0.6]{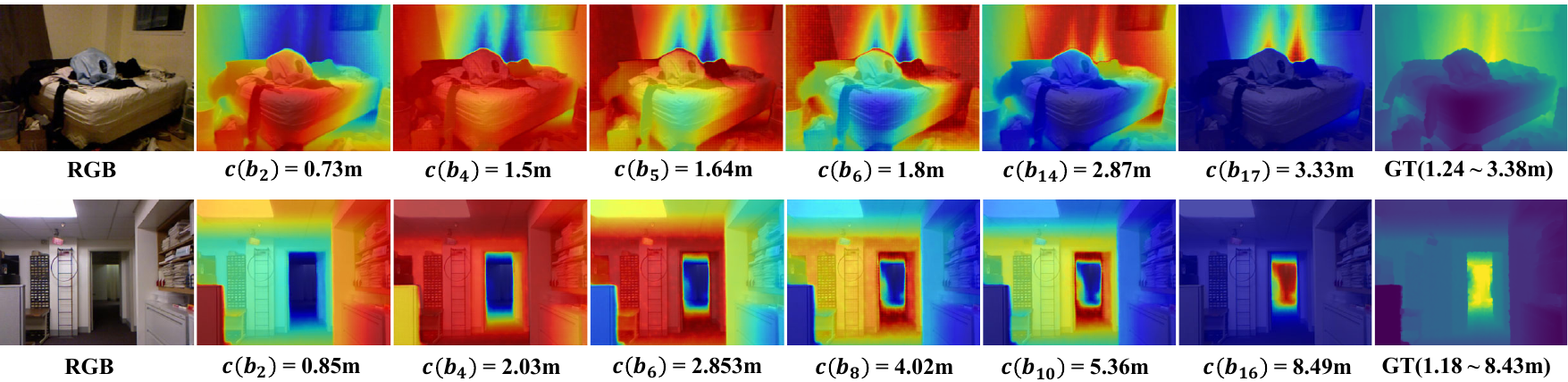}
    \caption{ 
    Visualization of probability representations corresponding to different bin centers. Red indicates higher probability, while blue indicates lower probability.
    }
    \label{fig:prob}
    \vspace{-10pt}
\end{figure*}

\begin{figure}[t]
    \centering
    \includegraphics[width=\columnwidth]{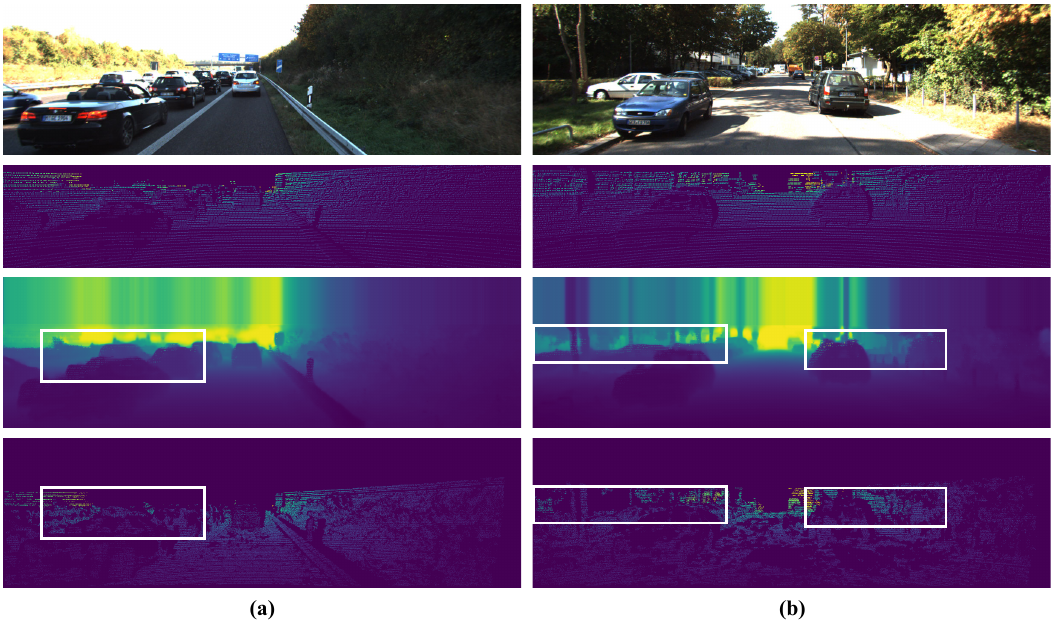}
    \caption{
    Our qualitative results on the KITTI dataset. From top to bottom are RGB images, the sparse depth maps, the result of our method, and the ground truth.
    }
    \label{fig:kitti}
\vspace{-10pt}
\end{figure}

\subsection{Qualitative Results}\label{sec:result_scannet}
\subsubsection{Comparison with SOTA}
As shown in Fig.~\ref{fig:nyu}, the proposed method produces high-quality depth maps with clear object contours on NYU-Depth-v2, even for tiny objects. For instance, in the first row, our model can perceive the very small depth difference between two refrigerator doors.
We show two examples of our prediction on KITTI in Fig.~\ref{fig:kitti}. There are multiple "discordant" depth points in our depth map, primarily due to the noise in the sparse depth map misleading our model's judgment of depth categories, and introducing inaccurate depth samples to the network.
The qualitative result on ScanNet-v2 is shown in Fig.~\ref{fig:scan}, our method can accurately predict the depth of small objects such as armrests, faucets and chair legs. Due to our progressive bi-directional information interactions manner with multi-scale supervision, the preceding stage features can provide more precise guidance for the subsequent stage, enabling our network to accurately identify these hard-to-detect small objects. 

\subsubsection{Multi-scale Depth Predictions}
Fig.~\ref{fig:multi_depth} shows predicted depth maps at different stages. We can observe that the quality of depth maps is improved with the increase of prediction stages which is more obvious in the first three stages. The improvement may benefit from two aspects.
The increased feature map resolution can represent more details and the bin partition with more depth categories contains more depth distribution priors.

\subsubsection{Bin Partitions} 
Results of Bin partitioning and corresponding scene depth distributions are illustrated in Fig.~\ref{fig:bins}. In Fig.~\ref{fig:bins} (a), pixels are clustered around small depth values which are mirrored in the bin partition. Conversely, in Fig.~\ref{fig:bins} (c), the depth distribution is more uniform. And our bin partition aligns with this distribution across continuous depth categories. This phenomenon endorses our motivation that the bin partition offers scene depth distribution priors.

\subsubsection{Probability Predictions} 
In Fig.~\ref{fig:prob}, we visualize the probability representation of some depth categories in the final stage.
As we can see, categories associated with small depth values yield high probabilities for regions close to the camera. With the increased depth values, attention shifts to distant areas, while close regions receive less attention. We attribute the rational probability representation to the interaction between depth decoupling and modulating branches, where the latter's predictions reference the former's bin partitions.

\renewcommand\arraystretch{1.2}
\begin{table}[b]
\small
\setlength{\abovecaptionskip}{0.01cm}
\setlength{\belowcaptionskip}{0.01cm}
\centering
\caption{The results of generalization experiments on NYU-Depth-v2. We report the performance of our method test under various sampling methods on the top. And report the performance of our method which is training without depth sample on the bottom. * means test on 5 depth samples.}
\resizebox{\columnwidth}{!}{
\begin{tabular}{c|cc|ccc}
\toprule[2pt]
& \multicolumn{2}{c|}{\textbf{Lower is better}} & \multicolumn{3}{c}{\textbf{Higher is better}}\\
\cline{2-6}
\multirow{-2}{*}{\textbf{Methods}} & AbsRel$\downarrow$ & RMSE$\downarrow$ & $\delta_{1.25}$ $\uparrow$& $\delta_{1.25}^2$ $\uparrow$& $\delta_{1.25}^3$ $\uparrow$ \\
\cline{1-6}
Random & 0.013 & 0.095 & 0.995 & 0.999 & 0.999 \\
\cline{2-6}
Grid  & \begin{tabular}[c]{@{}c@{}}0.014 \\ 
(\gr{-0.001})\end{tabular} & \begin{tabular}[c]{@{}c@{}}0.098 \\
(\gr{-0.003})\end{tabular} & \begin{tabular}[c]{@{}c@{}}0.995\\
({-0.000})\end{tabular} & \begin{tabular}[c]{@{}c@{}}0.999\\
({-0.000})\end{tabular} & \begin{tabular}[c]{@{}c@{}}0.999\\
(-0.000)\end{tabular} \\
\cline{2-6}
Top-bias & \begin{tabular}[c]{@{}c@{}}0.014\\
(\gr{-0.001})\end{tabular} & \begin{tabular}[c]{@{}c@{}}0.099\\
(\gr{-0.004})\end{tabular} & \begin{tabular}[c]{@{}c@{}}0.995\\
({-0.000})\end{tabular} & \begin{tabular}[c]{@{}c@{}}0.999\\
({-0.000})\end{tabular} & \begin{tabular}[c]{@{}c@{}}0.999\\
({-0.000})\end{tabular} \\
\cline{2-6}
Middle-bias & \begin{tabular}[c]{@{}c@{}}0.015\\
(\gr{-0.002})\end{tabular} & \begin{tabular}[c]{@{}c@{}}0.103\\
(\gr{-0.008})\end{tabular} & \begin{tabular}[c]{@{}c@{}}0.993\\
(\gr{-0.002})\end{tabular} & \begin{tabular}[c]{@{}c@{}}0.999\\
({-0.000})\end{tabular} & \begin{tabular}[c]{@{}c@{}}0.999\\
({-0.000})\end{tabular} \\
\cline{2-6}
Bottom-bias & \begin{tabular}[c]{@{}c@{}}0.014\\
(\gr{-0.001})\end{tabular} & \begin{tabular}[c]{@{}c@{}}0.100\\
(\gr{-0.005})\end{tabular} & \begin{tabular}[c]{@{}c@{}}0.995\\
({-0.000})\end{tabular} & \begin{tabular}[c]{@{}c@{}}0.999\\
({-0.000})\end{tabular} & \begin{tabular}[c]{@{}c@{}}0.999\\
({-0.000})\end{tabular} \\
\cline{2-6}
w/ noise & \begin{tabular}[c]{@{}c@{}}0.025\\
(\gr{-0.012})\end{tabular} & \begin{tabular}[c]{@{}c@{}}0.116\\
(\gr{-0.021})\end{tabular} & \begin{tabular}[c]{@{}c@{}}0.994\\
(\gr{-0.001})\end{tabular} & \begin{tabular}[c]{@{}c@{}}0.999\\
({-0.000})\end{tabular} & \begin{tabular}[c]{@{}c@{}}0.999\\
({-0.000})\end{tabular} \\
\cline{2-6}
w/o sample & \begin{tabular}[c]{@{}c@{}}0.678\\
(\gr{-0.665})\end{tabular} & \begin{tabular}[c]{@{}c@{}}2.270\\
(\gr{-2.175})\end{tabular} & \begin{tabular}[c]{@{}c@{}}0.015\\
(\gr{-0.980})\end{tabular} & \begin{tabular}[c]{@{}c@{}}0.046\\
(\gr{-0.953})\end{tabular}  & \begin{tabular}[c]{@{}c@{}}0.107\\
(\gr{-0.892})\end{tabular} \\
\cline{1-6}
Sparsity~\cite{conti2023sparsity}* & 0.131 & 0.467 & - & - & - \\
\textbf{Ours} & \textbf{0.079} & \textbf{0.307} & 0.825 & 0.960 & 0.990 \\
\bottomrule[2pt]
\end{tabular}
}
\label{tab:sample}
\vspace{-10pt}
\end{table}

\subsubsection{Convergence of the model}
\begin{figure}[t]
    \centering
    \includegraphics[width=1\columnwidth]{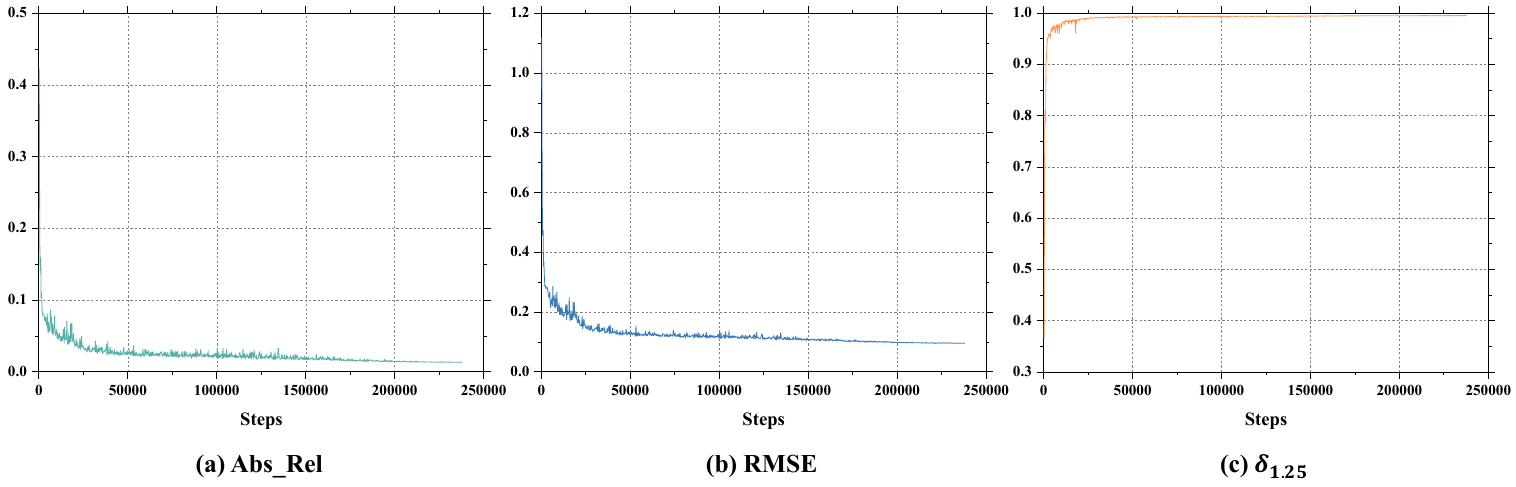}
    \caption{
    Convergence curves of metrics during the training progress on the NYU-Depth-v2 dataset.
    }
    \label{fig:curves}
    \vspace{-10pt}
\end{figure}

In Fig.~\ref{fig:curves}, we show the convergence curves of ``Abs Rel", ``RMSE" and ``$\delta_{1.25}$" metrics during the training progress. As we can see, the performance improves rapidly before 50,000 steps and then begins to converge gradually, which demonstrates our model can reach a quick and steady convergence. 
Thus our model can attain competitive performances and be trained easily with stable results.

\subsubsection{Point Cloud Visualizations} 
We transform the depths predicted by our methods into point clouds. As shown in Fig.~\ref{fig:point}, our method can capture the depth distribution and complete the depth accurately. Especially in the bottom line, the depth prediction of our method matches the depth distribution on the surface of the wall better.

\begin{figure}[t]
    \centering
    \includegraphics[width=1\columnwidth]{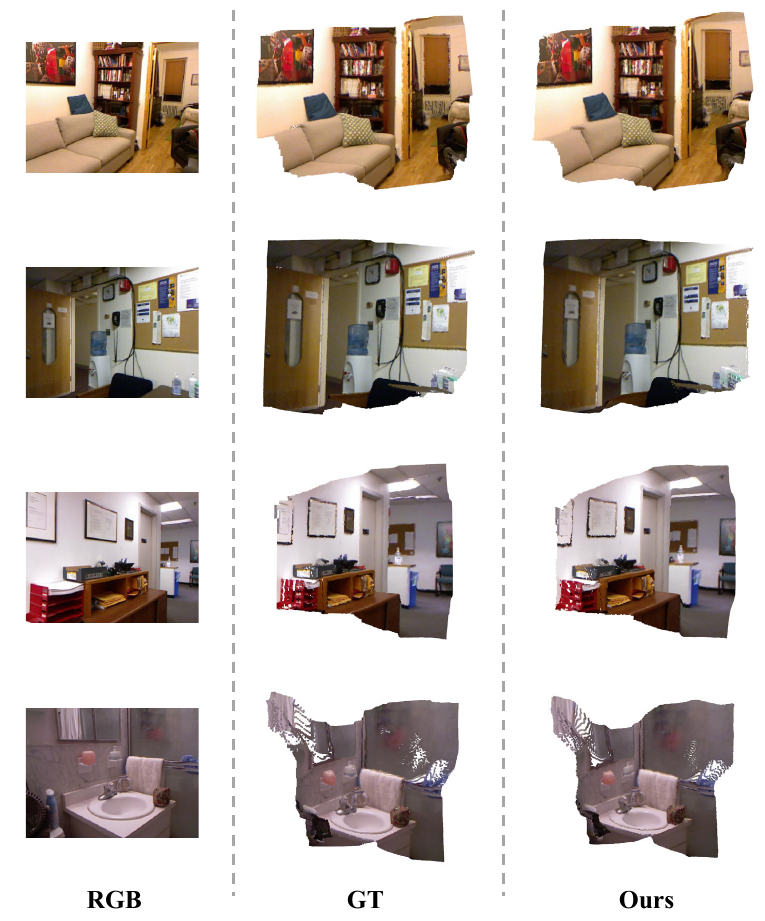}
    \caption{
    Visualization comparison of the point clouds generated by our method with the ground truth.
    }
    \label{fig:point}
    \vspace{-10pt}
\end{figure}

\section{Generalization}
To verify the generalization of our model under different sampling patterns, we adopt various depth sampling patterns to simulate sparse depth maps obtainable by models in practical applications. These sampling patterns include grid sampling, bottom-biased sampling, middle-biased sampling, top-biased sampling, sampling with noise, and a pattern with no available depth sampling points. Visualization illustrations of these sampling patterns are shown in Fig.~\ref{fig:point}. 

Specifically, the probability $p_{ij}$ of each pixel being sampled under all biased sampling patterns is given by:
\begin{align}
p_{i j}=\frac{1}{(D[i, j])^\alpha+1}.
\end{align}
Where $i$ and $j$ denote the row and column indices of pixels in the image respectively. $D$ represents the ground truth depth map. And $alpha$ serves as a smoothing factor to alleviate probability discrepancies between pixels, we set it to 0.35 to achieve better simulation results. Under the bottom-biased sampling pattern, $D[i, j]=i$. Under the top-biased sampling pattern, $D[i, j]=height-i$. Under the middle-biased sampling pattern, $D[i, j]= \mid i- height/2 \mid$. Then, we normalize these probabilities to ensure their sum equals 1. We employ Gaussian noise to simulate the noise generated when depth sensors produce inaccurate measurements, and each sample point is subject to a $50\%$ probability of being corrupted by noise.

\begin{figure}[t]
    \centering
    \includegraphics[width=1\columnwidth]{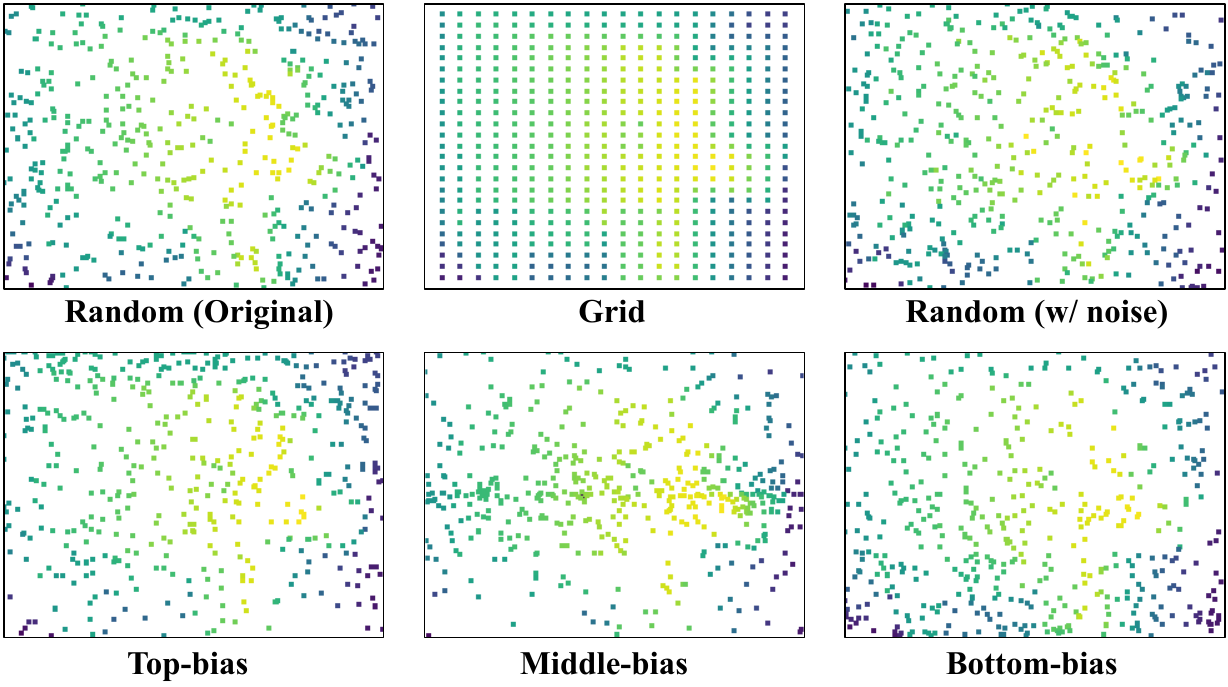}
    \caption{
    Visualization examples of sparse depth maps under different sampling patterns.
    }
    \label{fig:point}
\vspace{-10pt}
\end{figure}

We train the model under the original random sampling pattern and directly test it under these sampling patterns. As demonstrated in Table.~\ref{tab:sample}, our model exhibits slight performance degradation under grid sampling patterns, indicating its adaptability to this unbiased depth sampling pattern. Our model experiences performance degradation under the middle-biased sampling pattern while other patterns exhibit only slight performance loss, suggesting that regionally biased depth sampling has minimal impact on our model. This is because although sparse depth pays more attention to certain regions, it still partially reflects the depth distribution information of the scene. our model decreases to a certain degree under noise-affected sampling patterns which is primarily due to the inability of sparse depth samples to provide accurate depth distribution information under noise interference. Consequently, the depth distribution reflected by these sparse samples differs from the depth distribution learned by the model, leading to inaccurate depth category generation.
Finally, our model is significantly impacted by the pattern without available depth samples. This is because our bins initialization module fails to obtain coordinate information of valid samples in sparse depth maps, thereby it is unable to acquire scene depth distribution information, leading to erroneous initialization of depth categories. 

Additionally, even depth completion methods optimized for robustness to depth sample sparsity, such as Conti et al.~\cite{conti2023sparsity} experience a decrease of RMSE metric from 0.114 to 0.467 when only five depth sample points are available, which is only comparable to early monocular depth estimation methods. Thus, under conditions of extreme sparsity, we propose a solution involving a variant of the model that operates without sparse depth maps and subsequently report the result of our model variant in the last row.

\begin{figure}[t]
    \centering
    \includegraphics[width=1\columnwidth]{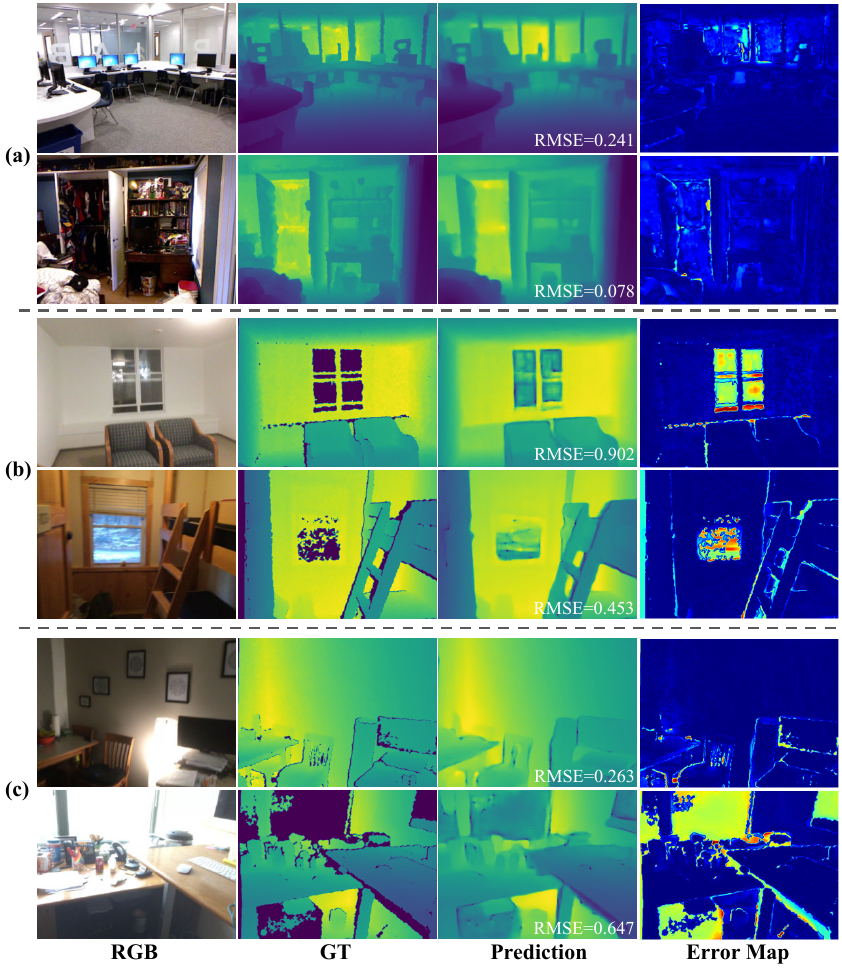}
    \caption{
    Failure examples visualization of our method which are complex scenes (a), scenes with windows (b), and scenes with strong lighting (c). 
    }
    \label{fig:fail}
\vspace{-10pt}
\end{figure}

\section{Limitation}
Although our approach has achieved competitive performance, shortcomings still exist in certain scenarios. As illustrated in Fig.~\ref{fig:fail}, we categorize the failure cases into three types, including complex scenes (a), scenes with windows (b), and scenes with strong lighting (c). For complex scenes, the model's misjudgment of the depth of such objects is attributed to the high quantity of objects and their similar textures. For scenes with windows, the model's inability to determine the depth values of corresponding pixels is due to the perspective and reflection characteristics of the windows. In the last scenario, texture loss caused by strong lighting leads to misjudgments of object edges or overall depth by the model. These three types of scenarios are difficult to address solely relying on RGB images. However, collecting sparse depth samples in corresponding regions could alleviate these issues, which underscores the significant role of sparse depth maps in depth completion tasks. The utilization of an adaptive depth measurement method~\cite{bergman2020deep, gofer2021adaptive, tcenov2022guide, dai2022adaptive, shomer2023prior} can be employed to enable depth sensors to capture areas with high model uncertainty.

\section{Conclusion}
In this paper, we develop a progressive depth decoupling and modulating algorithm for depth completion, which incrementally decouples the depth range into bins and adaptively generates multi-scale dense depth maps with corresponding probabilities. Specifically, we first design a bins initialization module to capture depth distribution information from sparse depth map for guiding depth discretization in different scenes. Second, we propose the progressive refinement strategy with bi-directional information interactions between depth decoupling and modulating branches and produce multi-scale depth maps. Finally, a multi-scale supervision loss is proposed to generate accurate guidance for subsequent stages and produce an accurate final depth prediction result. Extensive experiments on public datasets demonstrate the superiority of the proposed method. We find that leveraging distribution priors extracted from sparse depth maps is more beneficial to accurate depth prediction. Additionally, we demonstrate that progressive depth decoupling effectively alleviates the challenge of directly generating precise depth categories. Furthermore, we validate that incorporating intermediate supervision into multi-level prediction networks is advantageous for depth completion. Currently, our method still relies on the quality of sparse depth maps, and we plan to further investigate depth discretization methods robust to sparsity in depth samples in the future.

\bibliographystyle{IEEEtran}
\bibliography{ref}

\appendix

\renewcommand\arraystretch{1.2}
\begin{table*}[h]
\small
\centering
\caption{The detailed architecture parameters of our framework. All parameters are computed under the condition of bins number=64}
\resizebox{\textwidth}{!}{
\begin{tabular}{c|c|c|c|c|c}
\toprule[2pt]
\multicolumn{2}{c|}{Name} &Parameters &Input &Output &Description\\
\cline{1-6}
\multicolumn{2}{c|}{Encoder} &23.637M &$H\times W\times 4$ &$H/16\times W/16\times 512$ &\begin{tabular}[c]{@{}c@{}}ResNet34\cite{he2016deep}\\ 
Convolutional Layer (kernel\_size: $3\times 3$, stride: $1$, padding: $1$)\end{tabular}\\
\cline{1-6}
\multicolumn{2}{c|}{BIM} &0.014M &$H\times W\times 1$ &$m_1\times 128$  &\begin{tabular}[c]{@{}c@{}}Linear Layer input\_channel: $3$, output\_channel: $64$)\\ 
Linear Layer (input\_channel: $64$, output\_channel: $125$)\\ 
Convolutional Layer (kernel\_size: $3\times 3$, stride: $1$, padding: $1$)\end{tabular}\\
\cline{1-6}
\multirow{11}{*}{Decoder} 
    &\multirow{2}{*}{Block1} &\multirow{2}{*}{1.770M} &$H/16\times W/16\times 512$ &$H/8\times W/8\times 256$ &\multirow{2}{*}{\begin{tabular}[c]{@{}c@{}}Transpose Convolutional Layer (kernel\_size: $3$, stride: $2$, padding: $1$, output\_padding: $1$)\\ 
    Convolutional Layer (kernel\_size: $1\times 1$, stride: $1$, padding: 0)\end{tabular}}\\ 
    & & & $H/16\times W/16\times 128$ &$H/8\times W/8\times (m_1+2)$ &\\
    \cline{2-6}
    &\multirow{3}{*}{Block2} &\multirow{3}{*}{1.180M} &$H/8\times W/8\times 256$ &\multirow{3}{*}{\begin{tabular}[c]{@{}c@{}}$H/4\times W/4\times 128$\\
    $H/4\times W/4\times (m_2+2)$\end{tabular}} & \multirow{3}{*}{\begin{tabular}[c]{@{}c@{}}Transpose Convolutional Layer (kernel\_size: $3$, stride: $2$, padding: $1$, output\_padding: $1$)\\ 
    Convolutional Layer (kernel\_size: $3\times 3$, stride: $1$, padding: $1$)\\
    Convolutional Layer (kernel\_size: $1\times 1$, stride: $1$, padding: $0$)\end{tabular}} \\
    & & & $H/8\times W/8\times 512$ & & \\
    & & & $H/8\times W/8\times 128$ & & \\
    \cline{2-6}
    & \multirow{3}{*}{Block3} & \multirow{3}{*}{0.370M} & $H/4\times W/4\times 128$ & \multirow{3}{*}{\begin{tabular}[c]{@{}c@{}}$H/2\times W/2\times 64$\\       
    $H/2\times W/2\times (m_3+2)$\end{tabular}} & \multirow{3}{*}{\begin{tabular}[c]{@{}c@{}}Transpose Convolutional Layer (kernel\_size: $3$, stride: $2$, padding: $1$, output\_padding: $1$)\\
    Convolutional Layer (kernel\_size: $3\times 3$, stride: $1$, padding: $1$)\\
    Convolutional Layer (kernel\_size: $1\times 1$, stride: $1$, padding: $0$)\end{tabular}} \\
    & & & $H/4\times W/4\times 256$ & & \\
    & & & $H/4\times W/4\times 128$ & & \\
    \cline{2-6}
    & \multirow{3}{*}{Block4} & \multirow{3}{*}{0.259M} & $H/2\times W/2\times 64$ & \multirow{3}{*}{\begin{tabular}[c]{@{}c@{}}$H\times W\times 64$\\
    $H\times W\times (m_4+2)$\end{tabular}} & \multirow{3}{*}{\begin{tabular}[c]{@{}c@{}}Transpose Convolutional Layer (kernel\_size: $3$, stride: $2$, padding: $1$, output\_padding: $1$)\\
    Convolutional Layer (kernel\_size: $3\times 3$, stride: $1$, padding: $1$)\\
    Convolutional Layer (kernel\_size: $1\times 1$, stride: $1$, padding: $0$)\end{tabular}} \\
    & & & $H/2\times W/2\times 128$ & & \\
    & & & $H/2\times W/2\times 128$ & & \\
\cline{1-6}
\multicolumn{2}{c|}{\multirow{3}{*}{PPB}} & \multirow{3}{*}{0.299M} & $H\times W\times 64$ & \multirow{3}{*}{$H\times W\times (m_5+2)$} & \multirow{3}{*}{\begin{tabular}[c]{@{}c@{}}Convolutional Layer (kernel\_size: $3\times 3$, stride: $1$, padding: $1$)\\
Convolutional Layer (kernel\_size: $1\times 1$, stride: $1$, padding: $0$)\end{tabular}} \\
\multicolumn{2}{c|}{} & & $H\times W\times 64$ & & \\
\multicolumn{2}{c|}{} & & $H\times W\times 128$ & & \\
\cline{1-6}
\multirow{12}{*}{Transformer} 
    & Block1 & 0.659M & \begin{tabular}[c]{@{}c@{}}$m_1\times 128$\\
    $H*W/256\times 128$\end{tabular} &\begin{tabular}[c]{@{}c@{}}$m_1\times 128$\\
    $m_1\times 1$\\
    $m_2\times 128$\end{tabular} & \begin{tabular}[c]{@{}c@{}}TransformerDecoderLayer (embedding\_dim: $128$, num\_head: $4$, dim\_feedforward: $2048$)\\
    Linear Layer (input\_channel: $m_1$, output\_channel: $m_2$)\\
    Linear Layer (input\_channel: $128$, output\_channel: $1$)\end{tabular} \\
    \cline{2-6}
    & Block2  & 0.660M  & \begin{tabular}[c]{@{}c@{}}$m_2\times 128$\\
    $H*W/64 \times 128$\end{tabular} &\begin{tabular}[c]{@{}c@{}}$m_2\times 128$\\
    $m_2\times 1$\\ 
    $m_3\times 128$\end{tabular}  &\begin{tabular}[c]{@{}c@{}}TransformerDecoderLayer (embedding\_dim: $128$, num\_head: $4$, dim\_feedforward: $2048$)\\
    Linear Layer (input\_channel: $m_2$, output\_channel: $m_3$)\\
    Linear Layer (input\_channel: $128$, output\_channel: $1$)\end{tabular}  \\
    \cline{2-6}
    & Block3 & 0.660M & \begin{tabular}[c]{@{}c@{}}$m_3\times 128$\\ 
    $H*W/16 \times 128$\end{tabular} &\begin{tabular}[c]{@{}c@{}}$m_3\times 128$\\ 
    $m_3\times 1$\\
    $m_4\times 128$\end{tabular} &\begin{tabular}[c]{@{}c@{}}TransformerDecoderLayer   (embedding\_dim: $128$, num\_head: $4$, dim\_feedforward: $2048$)\\ 
    Linear Layer (input\_channel: $m_3$, output\_channel: $m_4$)\\ 
    Linear Layer (input\_channel: $128$, output\_channel: $1$)\end{tabular}\\
    \cline{2-6}
    & Block4 & 0.660M & \begin{tabular}[c]{@{}c@{}}$m_4\times 128$\\ 
    $H*W/4 \times 128$\end{tabular} & \begin{tabular}[c]{@{}c@{}}$m_4\times 128$\\ 
    $m_4\times 1$\\      
    $m_5\times 128$\end{tabular} &\begin{tabular}[c]{@{}c@{}}TransformerDecoderLayer   (embedding\_dim: $128$, num\_head: $4$, dim\_feedforward: $2048$)\\ 
    Linear Layer (input\_channel: $m_4$, output\_channel: $m_5$)\\ 
    Linear Layer (input\_channel: $128$, output\_channel: $1$)\end{tabular}\\
    \cline{2-6}
    & Block5 & 0.659M & \begin{tabular}[c]{@{}c@{}}$m_5\times 128$ \\
    $H*W \times 128$\end{tabular} & \begin{tabular}[c]{@{}c@{}}$m_5\times 128$ \\      
    $m_5\times 1$\end{tabular} & \begin{tabular}[c]{@{}c@{}}TransformerDecoderLayer   (embedding\_dim: $128$, num\_head: $4$, dim\_feedforward: $2048$) \\
    Linear Layer (input\_channel: $128$, output\_channel: $1$)\end{tabular} \\
\cline{1-6}
\multirow{5}{*}{Projection} 
    & 1 & 0.066M & $H/16\times W/16\times 512$ & $H*W/256 \times 128$ & Linear Layer (input\_channel: $512$, output\_channel: $128$) \\
    \cline{2-6}
    & 2 & 0.033M & $H/8\times W/8\times 256$ & $H*W/64\times 128$ & Linear Layer (input\_channel: $256$, output\_channel: $128$) \\
    \cline{2-6}
    & 3 & 0.017M & $H/4\times W/4\times 128$ & $H*W/16\times 128$ & Linear Layer (input\_channel: $128$, output\_channel: $128$) \\
    \cline{2-6}
    & 4 & 0.009M & $H/2\times W/2\times 64$ & $H*W/4\times 128$ & Linear Layer (input\_channel: $64$, output\_channel: $128$) \\
    \cline{2-6}
    & 5 & 0.009M & $H\times W\times 64$ & $H*W\times 128$ & Linear Layer (input\_channel: $64$,   output\_channel: $128$) \\
\cline{1-6}
\multirow{5}{*}{Inverse Projection} 
    & 1 & 0.001M & $m_1\times 128$ & $H/16\times W/16\times 128$ & \multirow{5}{*}{Convolutional Layer (kernel\_size: $1\times 1$, stride: $1$, padding: $0$)} \\
    \cline{2-5}
    & 2 & 0.010M & $m_2\times 128$ & $H/8\times W/8\times 128$ & \\
    \cline{2-5}
    & 3 & 0.075M & $m_3\times 128$ & $H/4\times W/4\times 128$ & \\
    \cline{2-5}
    & 4 & 0.572M & $m_4\times 128$ & $H/2\times W/2\times 128$ & \\
    \cline{2-5}
    & 5 & 4.505M & $m_5\times 128$ & $H\times W\times 128$ & \\
\cline{1-6}
\multicolumn{2}{c|}{CSPN Layer} & 0.005M & $H\times W\times (m_4+2)$ & $H\times W\times 8$ & Convolutional Layer (kernel\_size: $3\times 3$, stride: $1$, padding: $1$) \\
\bottomrule[2pt]
\end{tabular}
}
\label{tab1}
\end{table*}

\end{document}